\documentclass[journal]{IEEEtran}

\usepackage{epsfig}
\usepackage{graphicx}
\usepackage{amsmath}
\usepackage{amssymb}

\usepackage{xcolor}
\usepackage{soul}
\usepackage[utf8]{inputenc}
\usepackage[pagebackref=true,breaklinks=true,letterpaper=true,colorlinks,bookmarks=false]{hyperref}

\usepackage[figuresright]{rotating}
\usepackage{textcomp,subfigure,algorithm,algorithmic,multirow,upgreek,tabularx}
\usepackage{graphics,gensymb}
\usepackage{bm,cases,threeparttable}

\usepackage{setspace}

\graphicspath{{Figures/}}
\usepackage{booktabs} 
\newcommand{\tht}[2]{\begin{tabular}{@{}#1@{}}#2\end{tabular}}

\usepackage{amsfonts}
 
\usepackage{caption}
\usepackage{enumitem}
\setitemize{noitemsep,topsep=0pt,parsep=0pt,partopsep=0pt}

\ifCLASSINFOpdf
\else
\fi

\begin{document}
%
\title{Unified Generative Adversarial Networks for \\ Controllable Image-to-Image Translation}


\author{\IEEEauthorblockN{Hao Tang,
		Hong Liu, and
		Nicu Sebe
}
\thanks{Hao Tang and Nicu Sebe are with the Department of Information Engineering and Computer Science (DISI), University of Trento, Trento 38123, Italy. E-mail: hao.tang@unitn.it, sebe@disi.unitn.it.
\par Hong Liu is with the Shenzhen Graduate School, Peking University, Shenzhen 518055, China. E-mail: hongliu@pku.edu.cn.
\par Corresponding authors: Hao Tang and Hong Liu.}
}

\markboth{IEEE Transactions on Image Processing}%
{Shell \MakeLowercase{\textit{et al.}}: Bare Demo of IEEEtran.cls for IEEE Transactions on Magnetics Journals}
%

\IEEEtitleabstractindextext{%

\begin{abstract}
We propose a unified Generative Adversarial Network (GAN) for controllable image-to-image translation, i.e., transferring an image from a source to a target domain guided by controllable structures. In addition to conditioning on a reference image, we show how the model can generate images conditioned on controllable structures, e.g., class labels, object keypoints, human skeletons, and scene semantic maps.
The proposed model consists of a single generator and a discriminator taking a conditional image and the target controllable structure as input. 
In this way, the conditional image can provide appearance information and the controllable structure can provide the structure information for generating the target result. Moreover, our model learns the image-to-image mapping through three novel losses, i.e., color loss,  controllable structure guided cycle-consistency loss, and controllable structure guided self-content preserving loss. 
Also, we present the Fr\'echet ResNet Distance (FRD) to evaluate the quality of the generated images.
Experiments on two challenging image translation tasks, i.e., hand gesture-to-gesture translation and cross-view image translation, show that our model generates convincing results, and significantly outperforms other state-of-the-art methods on both tasks. 
Meanwhile, the proposed framework is a unified solution, thus it can be applied to solving other controllable structure guided image translation tasks such as landmark guided facial expression translation and keypoint guided person image generation.
To the best of our knowledge, we are the first to make
one GAN framework work on all such controllable structure guided image translation tasks. Code is available at \url{https://github.com/Ha0Tang/GestureGAN}.
\end{abstract}

}

\maketitle

\IEEEdisplaynontitleabstractindextext

%
\IEEEpeerreviewmaketitle

\section{Introduction}

\begin{figure*}[!t] \small
	\centering
	\includegraphics[width=1\linewidth]{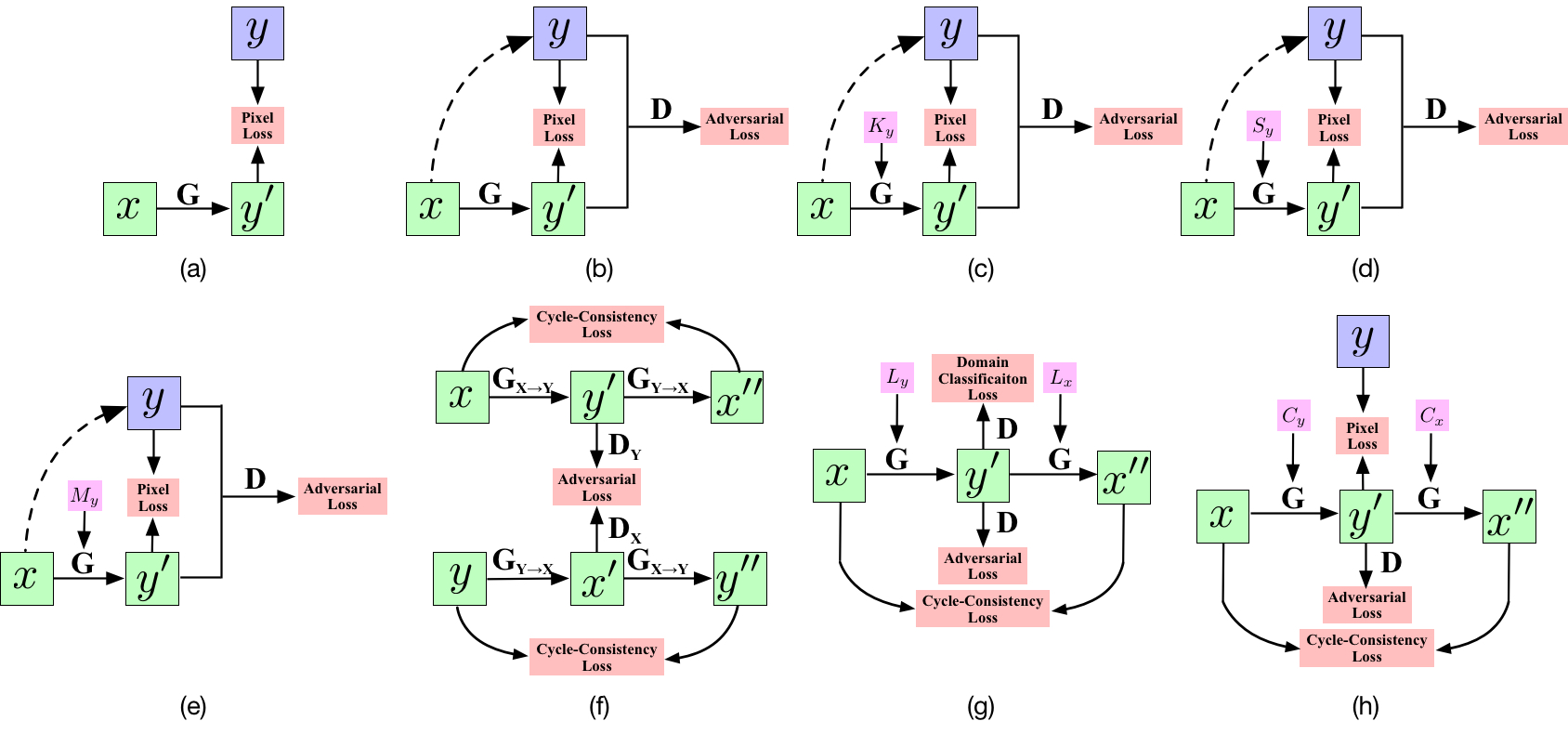}
	\caption{Comparison with state-of-the-art image-to-image translation methods. (a) Traditional deep learning methods, e.g., Context Encoder \cite{pathak2016context}. (b) Adversarial learning methods, e.g., Pix2pix~\cite{isola2017image} and BicycleGAN \cite{zhu2017toward}. (c) Keypoint-guided image generation methods, e.g., PG$^2$~\cite{ma2017pose}, G2GAN~\cite{song2017geometry} and DPIG~\cite{ma2017disentangled}. (d) Skeleton-guided image generation methods, e.g., SAMG~\cite{yan2017skeleton} and PoseGAN \cite{siarohin2017deformable}. (e) Semantic-guided image generation methods, e.g., SelectionGAN~\cite{tang2019multi} and X-Fork~\cite{regmi2018cross}. (f) Adversarial unsupervised learning methods, e.g., CycleGAN \cite{zhu2017unpaired}, DiscoGAN \cite{kim2017learning} and DualGAN \cite{yi2017dualgan}. (g) Multi-domain image translation methods, e.g., ComboGAN \cite{anoosheh2017combogan}, G$^2$GAN~\cite{tang2019dual} and StarGAN~\cite{choi2017stargan}. (h) Proposed GAN model in this paper. Note that the proposed GAN model is a unified solution for controllable structure guided image-to-image translation problem, i.e., controllable structure $C$ can be one of class label $L$, object keypoint $K$, human skeleton $S$ or semantic map $M$. 
		Notations: $x$ and $y$ are the real images; $x'$ and $y'$ are the generated images;  $x''$ and $y''$ are the reconstructed images; $K_y$ is the keypoint of $y$; $S_y$ is the skeleton of $y$; $M_y$ is the semantic map of $y$; $L_x$ and $L_y$ are the class labels of $x$ and $y$, respectively; $C_x$ and $C_y$ are the controllable structures of $x$ and $y$, respectively; $G$, $G_{X\mapsto Y}$ and $G_{Y\mapsto X}$ represent generators; $D$, $D_Y$ and $D_X$ denote discriminators.}
	\label{fig:framework_comparsion}
	\vspace{-0.4cm}
\end{figure*}

\IEEEPARstart{G}{enerative} Adversarial Networks (GANs) \cite{goodfellow2014generative} are generative models based on game theory, which have achieved impressive performance in a wide range of applications such as high-quality image generation~\cite{karras2018style,brock2018large}.
To generate specific kinds of images, Mirza et al.~\cite{mirza2014conditional} propose Conditional GANs (CGANs), which comprise a vanilla GAN model and other external controllable structure information, such as class labels~\cite{choi2017stargan},
reference images~\cite{isola2017image}, object keypoints~\cite{reed2016learning,ma2017pose}, human skeletons \cite{tang2018gesturegan,siarohin2017deformable}, and semantic maps~\cite{tang2019multi,park2019semantic,liu2020exocentric}

In this paper, we mainly focus on the image-to-image translation task using CGANs.
At a high level, current image-to-image translation techniques usually fall into one of two types:
supervised/paired \cite{isola2017image,wang2018high} and unsupervised/unpaired \cite{zhu2017unpaired,choi2017stargan}.
However, existing image-to-image translation frameworks are inefficient for the multi-domain image-to-image translation task. 
For instance, given $n$ different image domains, Pix2pix~\cite{isola2017image} and BicycleGAN~\cite{zhu2017toward} need to train $A_n^2{=}n(n{-}1){=}\Theta(n^2)$ models.
CycleGAN~\cite{zhu2017unpaired}, DiscoGAN~\cite{kim2017learning} and DualGAN~\cite{yi2017dualgan} need to train $C_n^2{=}\frac{n(n{-}1)}{2}{=}\Theta(n^2)$ models, or $n(n{-}1)$ generator/discriminator pairs since one model has two different generator/discriminator pairs for these methods.
ComboGAN~\cite{anoosheh2017combogan} requires $n{=}\Theta(n)$ models.
G$^2$GAN~\cite{tang2019dual} needs to train two generators, i.e., the generation generator and the reconstruction generator, while StarGAN \cite{choi2017stargan} only needs one model. 
However, for some specific image-to-image translation applications such as hand gesture-to-gesture translation~\cite{tang2018gesturegan} and person image generation~\cite{ma2017pose,tang2020bipartite}, $n$ could be arbitrary large since hand gestures and human bodies in the wild can have arbitrary poses, sizes, appearances, locations, and self-occlusions.
 
To address these limitations, several works have been proposed to generate images based on controllable structures, e.g., object keypoints, human skeleton, and scene semantic maps.
These works can be divided into three different categories:  
1) Object keypoint guide methods. Reed et al.~\cite{reed2016learning} proposed GAWWN, which generates bird images conditioned on bird keypoints.
Song et al.~\cite{song2017geometry} propose G2GAN for facial expression synthesis based on facial landmarks.
Ma et al. propose PG$^2$~\cite{ma2017pose} and a two-stage reconstruction pipeline~\cite{ma2017disentangled} achieving person image translation using a conditional image and a target pose image.
2) Human skeleton guided methods. Siarohin et al. \cite{siarohin2017deformable} introduce PoseGAN based on the human skeleton for human image generation.
Tang et al.~\cite{tang2018gesturegan} propose a novel GestureGAN conditioned hand skeleton for hand gesture-to-gesture image translation tasks.
Yan et al.~\cite{yan2017skeleton} propose a method to generate human motion sequences with simple backgrounds using CGANs and human skeleton information.
3) Scene semantic guide methods.
Wang et al.~\cite{wang2018high} propose Pix2pixHD, which can be used for turning semantic label maps into photo-realistic images or synthesizing portraits from face label maps.
Park et al.~\cite{park2019semantic} propose the spatially-adaptive normalization, a simple but effective layer for synthesizing images given an input semantic layout.
Regmi and Borji~\cite{regmi2018cross} propose X-Fork and X-Seq, which aim to generate images across two drastically different views by using the guidance of semantic maps.

The aforementioned methods have achieved impressive results on the corresponding tasks.
However, each of them is tailored for a specific application limiting their capability to generalize. Our framework does not impose any  application-specific constraint. This makes our setup considerably simpler than the other approaches (see Fig.~\ref{fig:framework_comparsion}).
To achieve this goal, we propose a unified solution for controllable image-to-image translation.
It allows generating high-quality images with arbitrary poses, sizes, structures, and locations in the wild.
Our GAN model only consists of one generator and one discriminator, taking  a conditional  image and the novel controllable structures as inputs.
In this way, the conditional image can provide appearance information and the controllable structures can provide structure information for generating the target image. 
In addition, to better learn the mapping between inputs and outputs, we propose three novel losses, i.e., color loss,  controllable structure guided cycle-consistency loss, and self-content preserving loss.
The proposed color loss can handle the problem of `channel pollution' that is frequently occurring in generative models such as PG$^2$ \cite{ma2017pose}, leading the generated images to be sharper and having higher quality.
The proposed controllable structure guided cycle-consistency loss is more flexible than the one proposed in CycleGAN~\cite{zhu2017unpaired}, reducing further the space of possible mappings between different domains.
The proposed self-content preserving loss can preserve color composition, object identity, and global layout of generated images.
These optimization loss functions and the proposed GAN framework are jointly trained in an end-to-end fashion to improve both fidelity and visual naturalness of the generated images.
Furthermore, we propose the Fr\'echet ResNet Distance (FRD), which is a novel and better evaluation metric to evaluate the generated images of GANs.
Extensive experiments on two challenging controllable image-to-image translation tasks with four different datasets, i.e., hand gesture-to-gesture translation and cross-view image translation, demonstrate that the proposed GAN model generates high-quality images with convincing details and achieves state-of-the-art performance on both tasks. 
Finally, the proposed GAN model is a general-purpose solution that can be applied to solve a wide variety of controllable structure guided image-to-image translation problems.

In summary, the contributions of this paper are as follows:
\begin{itemize}[leftmargin=*]
\item We propose a unified GAN model for controllable image-to-image translation tasks, which can generate target images with arbitrary poses, sizes, structures, and locations in the wild. 
\item We propose three novel objective functions to better optimize the proposed GAN model, i.e.,  color loss, controllable structure guided cycle-consistency loss, and self-content preserving loss. 
These optimization functions and the proposed GAN framework are jointly trained in an end-to-end fashion to improve both the quality and fidelity of the generated images.
\item We propose an efficient Fr\'echet ResNet Distance (FRD) metric to evaluate the similarity of the real and generated images, which is more consistent with human judgment. 
\item  Qualitative and quantitative results demonstrate the superiority of the proposed GAN model over the state-of-the-art methods on two challenging controllable image translation tasks with four datasets, i.e., hand gesture-to-gesture translation and cross-view image translation. 
\end{itemize}

Part of this work has been published in \cite{tang2018gesturegan}. 
We extend it in numerous ways: 
1) We extend GestureGAN proposed in \cite{tang2018gesturegan} to a unified GAN framework for handling all controllable image-to-image translation tasks.
2) We further tune our whole pipeline and improve its performance and generalizability for hand gesture-to-gesture translation and cross-view image translation by employing three additional losses, i.e., controllable structure guided self-content preserving loss, perceptual loss, and Total Variation loss. Moreover, we extend the one-cycle framework in \cite{tang2018gesturegan} to a two-cycle framework and validate the effectiveness. 
3) We extend the experimental evaluation provided in \cite{tang2018gesturegan} in several directions. First, we conduct extensive experiments on two challenging generative tasks with four different datasets, demonstrating the wide application scope of our GAN framework. Second, we conduct exhaustive ablation studies to evaluate each component of the proposed method. Third, we investigate the influence of hyper-parameters on generation performance. Forth, we compare the model parameters of different methods. Lastly, we provide arbitrary image translation results on both tasks.
\section{Related Work}
\label{sec:relatewprk}

\noindent\textbf{Generative Adversarial Networks (GANs)} are unsupervised learning methods and have been proposed in~\cite{goodfellow2014generative}.
Recently, GANs have shown promising results in various applications, e.g., image generation \cite{karras2018style,brock2018large,zhang2020dual}.
Existing approaches employ the idea of GANs for conditional image generation, such as image-to-image translation~\cite{isola2017image}, text-to-image translation~\cite{qiao2019mirrorgan,tao2020df}, audio-to-image \cite{duan2019cascade}, and sketch generation~\cite{tang2019attribute,chen2018sketchygan}. 
The key success of GANs is the adversarial loss, which allows the model to generate images that are  indistinguishable from real images, and this is exactly the goal that many tasks aim to optimize.
In this paper, we mainly focus on image-to-image translation tasks.

\noindent\textbf{Image-to-Image Translation} is the problem of transferring an image from a source domain to a target domain, which uses input-output data to learn a parametric mapping between inputs and outputs, e.g., Isola et al. \cite{isola2017image} propose Pix2pix, which uses a conditional GAN to learn a translation function from input to output image domain with paired training data.
However, collecting large sets of image pairs is often prohibitively expensive or unfeasible.
To solve this limitation, Zhu et al.~\cite{zhu2017unpaired} propose CycleGAN, which can learn to translate between domains without paired input-output examples by using the cycle-consistency loss.
Similar ideas have been proposed in several works \cite{yi2017dualgan,choi2017stargan,tang2019dual}.
For example, 
Choi et al.~\cite{choi2017stargan} introduce StarGAN, which can perform image-to-image translation for multiple domains. 

However, existing image-to-image translation models are inefficient and ineffective.
For example, with $n$ image domains, CycleGAN~\cite{zhu2017unpaired}, DiscoGAN~\cite{kim2017learning}, and DualGAN~\cite{yi2017dualgan} need to train $2C_n^2{=}n(n{-}1){=}\Theta(n^2)$ generators and discriminators, while Pix2pix \cite{isola2017image} and BicycleGAN \cite{zhu2017toward} have to train $A_n^2{=}n(n{-}1){=}\Theta(n^2)$ generator/discriminator pairs.
Recently, Anoosheh et al. propose ComboGAN~\cite{anoosheh2017combogan}, which only needs to train $n$ generator/discriminator pairs for $n$ different image domains, having a complexity of $\Theta(n)$.
Tang et al.~\cite{tang2019dual} propose G$^2$GAN, which can perform image-to-image translations for multiple domains using only two generators, i.e., the generation generator and the reconstruction generator.
Additionally, Choi et al.~\cite{choi2017stargan} propose StarGAN, in which a single generator and a discriminator can perform unpaired image-to-image translations for multiple domains.
Although the computational complexity of StarGAN is $\Theta(1)$, this model has only been validated on the face attributes modification task with clear background and face cropping.
More importantly, for some specific image-to-image translation tasks such as hand gesture-to-gesture translation~\cite{tang2018gesturegan} and person image generation~\cite{ma2017pose,tang2020bipartite} tasks, the image domains could be arbitrarily large, e.g., hand gestures and human bodies in the wild can have arbitrary poses, sizes, appearances, structures, locations, and self-occlusions.
The aforementioned approaches are not effective in solving these specific situations.

\begin{figure*}[!t] \small
	\centering
	\includegraphics[width=1\linewidth]{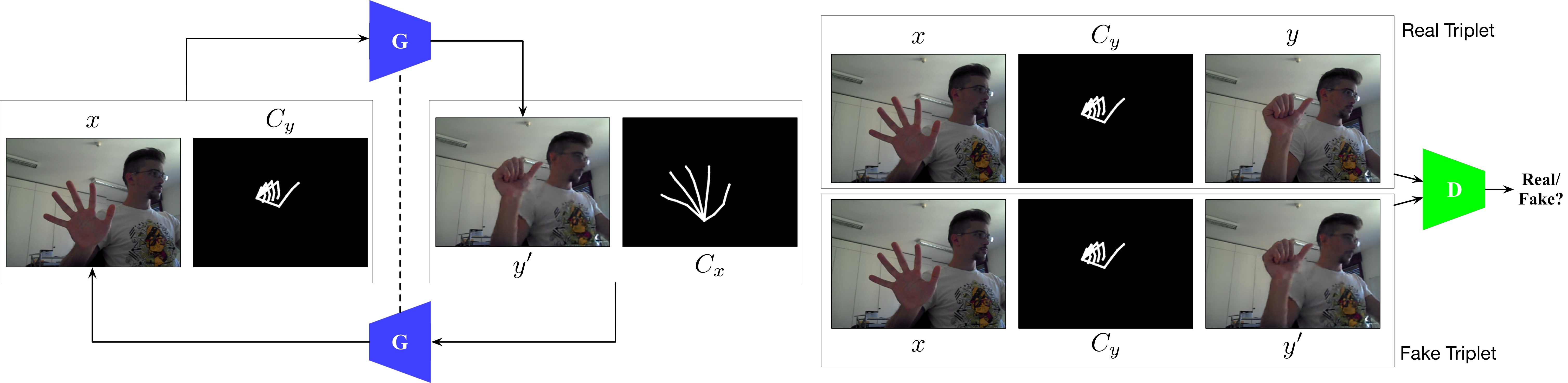}
	\caption{Pipeline of the proposed unified GAN model for controllable image-to-image translation tasks.
	The proposed GAN framework consists of a single generator~$G$ and an associated adversarial discriminator~$D$, which takes a conditional image $x$ and a controllable structure $C_y$ as input to produce the target image $y'$. We have two cycles and here we only show one of them. Note that the controllable structure $C_y$ can be class labels, object keypoints, human skeletons, semantic maps, etc.	
	}
	\label{fig:framework}
	\vspace{-0.4cm}
\end{figure*}

\noindent\textbf{Controllable Image-to-Image Translation.}
To fix these limitations, several recent works have been proposed to generate persons, birds, faces and scene images based on controllable structures, i.e., object keypoints~\cite{reed2016learning,ma2017pose,ma2017disentangled}, human skeletons \cite{yan2017skeleton,tang2019cycle,siarohin2017deformable,tang2020xinggan} and semantic maps~\cite{wang2018high,park2019semantic,regmi2018cross,tang2020local}.
In this way, controllable structures provide four types of information to guide the image generation process, i.e., category, scale, orientation, and location.
Although significant efforts have been made to achieve controllable image-to-image translation in the area of computer vision, there has been very limited research on universal controllable image translation.
That is, the typical problem with the aforementioned generative models is that each of them is tailored for a specific application, which greatly limits the generalization ability of the proposed models.
To handle this problem, we propose a novel and unified GAN model, which can be tailored for handling all kinds of problem settings of controllable structure guided image-to-image translation, including object keypoint guided generative tasks, human skeleton guided generative tasks and semantic map guided generative tasks, etc.
\section{Model Description}
\label{sec:method}

In this section, we present the details of the proposed GAN model (Fig.~\ref{fig:framework}).
We present a controllable structure guided generator, which utilizes the images from one domain and conditional controllable structures from another domain as inputs and produces images in the target domain.
Moreover, we propose a novel discriminator which also takes the controllable structure into consideration.
The proposed GAN model is trained in an end-to-end fashion mutually benefiting from the generator and the discriminator.

\subsection{Controllable Structure Guided Generator}
\noindent \textbf{Controllable Structure Guided Generation.} 
Image-to-image translation tasks, such as hand gesture-to-gesture translation~\cite{tang2018gesturegan}, person image generation~\cite{ma2017pose}, facial expression-to-expression translation~\cite{tang2019cycle} and cross-view image translation~\cite{regmi2018cross} are very challenging.
In these tasks, the source domain and the target domain may have large  deformations.
Moreover, these tasks can be treated as an infinite mapping problem leading to ambiguity issues in the translation process.
For instance, in the hand gesture-to-gesture translation task, if you input a hand gesture image to the generator, it has no idea which gestures should output.

To fix this limitation, we employ controllable structures as conditional guidance to guide the image generation process.
The controllable structures can be class labels, object keypoints, human skeletons or semantic maps, etc.
Following \cite{ma2017pose,siarohin2017deformable,regmi2018cross} we generate the controllable structures using deep models pretrained from other large-scale datasets, e.g., we apply the pose estimator OpenPose \cite{cao2017realtime} to obtain approximate human body poses and hand skeletons.
Specifically, as shown in Fig.~\ref{fig:framework}, we concatenate the input conditional image $x$ from the source domain and the controllable structure $C_y$ from a target domain, and input them into the generator $G$ and synthesize the target image $y'{=}G(x, C_y)$.
In this way, the ground-truth controllable structure $C_y$ provides stronger supervision and structure information to guide the image-to-image translation in the deep network, while the conditional image $x$ provides the appearance information to produce the final result $y'$.

\noindent \textbf{Controllable Structure Guided Cycle.} 
Guided by the controllable structure $C_y$, our generator can produce the corresponding image $y'$. 
However, state-of-the-art controllable image-to-image translation methods such as~\cite{ma2017pose,siarohin2017deformable,tang2019multi,regmi2018cross} only consider the image translation process, i.e., from the source domain to the target domain.
Different from them,  we consider both the image translation process and image reconstruction process, i.e., from the source domain to the target domain and from the target domain back to the source domain.
The  intuition  behind this is that if we translate from one domain to the other and back again we should arrive at where we started.
The proposed controllable structure guided cycle is different from the cycle proposed in CycleGAN~\cite{zhu2017unpaired}, which uses a cycle-consistency loss to preserve the content of its input images while changing only the domain-related part of the inputs.
The main difference is that CycleGAN can only handle two different domains, while an image translation problem such as hand gesture-to-gesture translation task has arbitrary domains, e.g., hand gestures in the wild can have arbitrary poses, sizes, appearances, structures, locations, and self-occlusions.
Therefore, we need the controllable structure to guide the learning of the proposed cycle.
The proposed controllable structure guided cycle is also different from the cycle proposed in StarGAN~\cite{choi2017stargan}, which translates an original image into an image in the target domain and then reconstructs the original image from the translated image through feeding the target label.
However, class labels can only provide the category information, while the controllable structure can provide four types of information for generation at the same time, i.e., category, location, scale, and orientation. 
Specifically, as shown in Fig.~\ref{fig:framework}, the generated image $y'$ and the controllable structure $C_x$ are concatenated to input into the generator~$G$.
Thus, the proposed controllable structure guided cycle can be formulated as follows,
\begin{equation}
\begin{aligned}
x''= G(y', C_x) = G(G(x, C_y), C_x)  \approx x.
\end{aligned}
\end{equation}
Note that we use a single generator twice, first to translate an original image into an image in the target domain and then to reconstruct the original image from the translated image.
Image translation and image reconstruction are simultaneously considered in our framework, constructing a full mapping cycle.
Similarly, we have another cycle,
\begin{equation}
\begin{aligned}
y''= G(x', C_y) = G(G(y, C_x), C_y)  \approx y.
\end{aligned}
\end{equation}

\noindent \textbf{Controllable Structure Guided Cycle-Consistency Loss.}
To better optimize the proposed cycle,  we propose a novel controllable structure guided cycle-consistency loss.
It is worth noting that CycleGAN~\cite{zhu2017unpaired} is different from the Pix2pix model~\cite{isola2017image} as the training data in CycleGAN is unpaired. 
CycleGAN introduces the cycle-consistency loss to enforce forward-backward consistency.
In that case, the cycle-consistency loss can be regarded as `pseudo' pairs of training data even though we do not have the corresponding data in the target domain which corresponds to the input data from the source domain.
However, in this paper, we introduce the controllable structure guided cycle-consistency loss for the paired image-to-image translation task.
This loss ensures the consistency between source images and the reconstructed image, and it can be expressed as,
\begin{equation}
\begin{aligned}
\mathcal{L}_{cyc}(G, C_x, C_y) = & \mathbb{E}_{x, C_x, C_y} \left[ \left|\left| x - G(G(x, Cy), C_x) \right|\right|_1 \right] \\
+ &  \mathbb{E}_{y, C_x, C_y} \left[ \left|\left| y - G(G(y, Cx), C_y) \right|\right|_1 \right], 
\end{aligned}
\label{eqn:con}
\end{equation}
where $G$ is the generator; $x$ and $y$ are the input images; $C_x$ and $C_y$ are the controllable structures of image $x$ and $y$,  respectively.
As mentioned before, we use the same generator~$G$ twice.
Equipped with this loss, the proposed generator~$G$ further improves the image quality due to its implicit data augmentation effect from a multi-task learning setting.

\subsection{Controllable Structure Guided Discriminator}

Conditional GANs (CGANs) such as Pix2pix~\cite{isola2017image} learn the mapping $G(x) \mapsto y$, where $x$ is the input conditional image.
Generator $G$ is trained to generate image $y'$ that cannot be distinguished from `real' image $y$ by an adversarially trained discriminator $D$, while the discriminator $D$ is trained as well as possible to detect the `fake' images generated by the generator $G$.  
The objective function of CGANs is defined as follows,
\begin{equation}\small
\begin{aligned}
& \mathcal{L}_{cGAN}(G, D) =  \mathbb{E}_{x, y} \left[ \log D(x, y) \right] + 
\mathbb{E}_{x} \left[\log (1 - D(x, G(x))) \right],
\end{aligned}
\label{eqn:conditonalgan}
\end{equation}
where generator $G$ tries to minimize this objective function while the discriminator $D$ tries to maximize it. 
Thus, the solution is $G^*{=}\arg \min\limits_G \max\limits_D \mathcal{L}_{cGAN}(G, D).$
In this paper, we try to learn two mappings through one generator, i.e., $G(x, C_y) {\mapsto} y$ and $G(y', C_x) {\mapsto} x$.
As shown in Fig.~\ref{fig:framework},  in order to learn both mappings, we employ the controllable structures explicitly.
Thus, the adversarial losses of the two mappings are defined respectively, as follows:
\begin{equation}\small
\begin{aligned}
\mathcal{L}_{adv}(G, D, C_y) =  & \mathbb{E}_{[x,C_y], y} \left[\log D([x, C_y], y) \right]   \\
+ & \mathbb{E}_{[x,C_y]} \left[\log (1 - D([x, C_y], G(x, C_y))) \right],
\end{aligned}
\label{eqn:keypointgan}
\end{equation}
where $C_y$  is the controllable structure of image $y$ and $[\cdot,\cdot]$ represents the concatenation operation. 
This controllable structure guided input encourages $D$ to capture the local-aware information and generate semantic-matched target images.
Similarly, we have another adversarial loss,
\begin{equation}\small
\begin{aligned}
\mathcal{L}_{adv}(G, D, C_x) = & \mathbb{E}_{[y,C_x], x} \left[\log D([y, C_x], x) \right]  \\
+  & \mathbb{E}_{[y,C_x]} \left[\log (1 - D([y, C_x], G(y, C_x))) \right].
\end{aligned}
\label{eqn:keypointgan1}
\end{equation}
Thus, the final adversarial loss is the sum of Eq.~\eqref{eqn:keypointgan} and~\eqref{eqn:keypointgan1},
\begin{equation}\small
\begin{aligned}
\mathcal{L}_{adv}(G, D) = \mathcal{L}_{adv}(G, D, C_x) + \mathcal{L}_{adv}(G, D, C_y).
\end{aligned}
\label{eqn:adv}
\end{equation}

\subsection{Optimization Objective}
\noindent \textbf{Color Loss.} Previous work indicates that mixing the adversarial loss with a traditional loss such as $L1$ loss~\cite{isola2017image} or $L2$ loss~\cite{pathak2016context} between the generated images and the ground truth images improves the generation performance.
The definitions of $L1$ and $L2$ losses are:
\begin{equation}
\begin{aligned}
\mathcal{L}_{L{\{1,2\}}}(G)  = & \mathbb{E}_{[x, C_y], y} \left[ \lVert y - G([x, C_y]) \lVert_{\{1,2\}} \right], \\
+ &  \mathbb{E}_{[y, C_x], x} \left[ \lVert x - G([y, C_x]) \lVert_{\{1,2\}} \right].
\end{aligned}
\label{equ:l2}
\end{equation}

\begin{figure}[!t]\small
	\centering
	\includegraphics[width=1\linewidth]{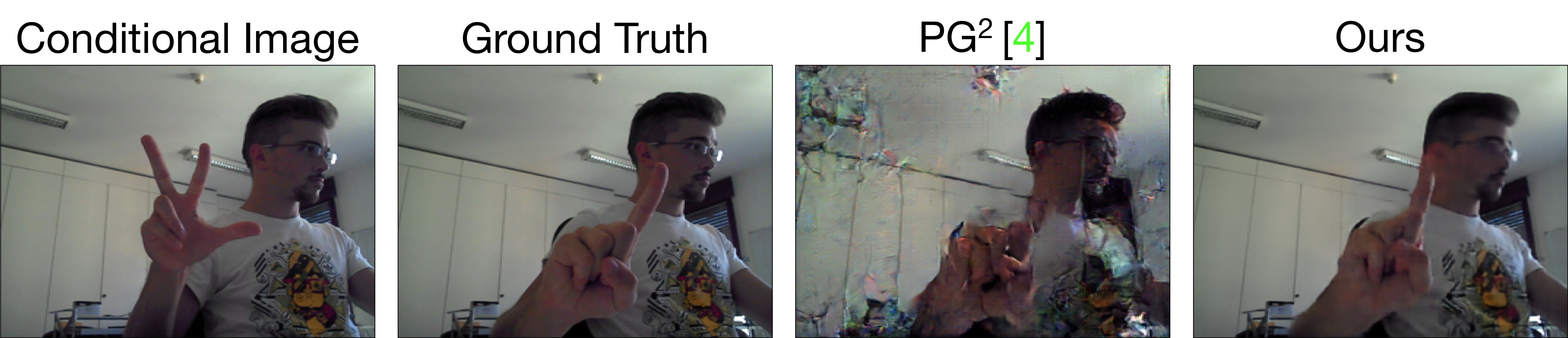}
	\caption{Illustration of the `channel pollution' issue. From left to right: Conditional Image, Ground Truth, PG$^2$~\cite{ma2017pose}, and Ours.
}
	\label{fig:first_figure}
	\vspace{-0.4cm}
\end{figure}

However, we observe that the existing image-to-image translation models such as PG$^2$~\cite{ma2017pose} cannot retain the holistic color of the input images.
An example is shown in Fig.~\ref{fig:first_figure}, where PG$^2$ is affected by the pollution issue and produces more unrealistic regions. 
Therefore, to address this limitation we introduce a novel channel-wise color loss.
Traditional generative models convert an entire image into another, which leads to the `channel pollution' problem.
However, the color loss treats $r$, $g$, and $b$ channels independently and generates only one channel each time, and then these three channels are combined to produce the final image.
Intuitively, since the generation of a three-channel image space is much more complex than the generation of single-channel image space, leading to a higher possibility of artifacts, we independently generate each channel.
The objective of $r$, $g$ and $b$ channel losses can be defined as follows,
\begin{equation}
\begin{aligned}
\mathcal{L}_{color_{c}{\{1,2\}}}(G)  = & \mathbb{E}_{[x_c, C_y], y_c} \left[ \lVert y_c - G([x_c, C_y]) \lVert_{\{1,2\}} \right] \\
 + & \mathbb{E}_{[y_c, C_x], x_c} \left[ \lVert x_c - G([y_c, C_x]) \lVert_{\{1,2\}} \right].
\end{aligned}
\label{eqn:color_rgb}
\end{equation}
where $c{\in}\{r,g,b\}$; $x_r$, $x_g$, and $x_b$ denote the $r$, $g$, and $b$ channel of image $x$, respectively, similar to $y_r$, $y_g$ and $y_b$; $\lVert\cdotp \lVert_1$ and $\lVert\cdotp \lVert_2$ represent $L1$ and $L2$ distance losses.
Thus, the color $L1$ and $L2$ losses can be expressed as, 
\begin{equation}
\begin{aligned}
\mathcal{L}_{color{\{1,2\}}}(G) = \mathcal{L}_{Color_{r}{\{1,2\}}} + \mathcal{L}_{Color_{g}{\{1,2\}}} + \mathcal{L}_{Color_{b}{\{1,2\}}}.
\end{aligned}
\label{eqn:color2}
\end{equation}
In Eq.~\eqref{equ:l2}, one channel is always influenced by the errors from other channels. 
On the contrary, if we compute the loss for each channel independently as shown in Eq.~\eqref{eqn:color2}, we can avoid such influence.
In this way, the error in one channel will not influence other channels.
We observe that this novel loss can improve the image quality in our experimental section. 

\noindent \textbf{Controllable Structure Guided Self-Content Preserving Loss.} To preserve the image content information (e.g., color composition, object identity, global layout) between the input and output, 
CycleGAN~\cite{zhu2017unpaired} proposes the identity preserving loss.
However, different from \cite{zhu2017unpaired}, we propose the controllable structure guided self-content preserving loss, which can be expressed as follows,
\begin{equation}
	\begin{aligned}
		 \mathcal{L}_{con}(G)  = & \mathbb{E}_{x, C_x} \left[ \left|\left| x - G(x, C_x) \right|\right|_1 \right] \\
		+ & \mathbb{E}_{y, C_y} \left[ \left|\left| y - G(y, C_y) \right|\right|_1 \right].
	\end{aligned}
	\label{eqn:preserve}
\end{equation}
We aim to minimize the $L1$ difference between the real image $x/y$ and the self-content preserving image $G(x, C_x)/G(y, C_y)$ for content information preservation. 
In this way, we regularize the generator to be near a self-content mapping when real images and self controllable structures are provided as the input to the generator.

\noindent \textbf{Perceptual Loss}
measures the perceptual similarity in a high-level feature space.
This loss has been shown to be useful for many tasks such as
style transfer~\cite{johnson2016perceptual} and image translation~\cite{wang2018high}.
The formulation of this loss is as follows:
\begin{equation}
\begin{aligned}
 \mathcal{L}_{vgg}(y') = \frac{1}{W_{i,j}H_{i,j}}\sum_{w=1}^{W_{i,j}}\sum_{h=1}^{H_{i,j}} \vert \vert \mathcal{F}^{k} (y)-\mathcal{F}^{k}(G(x, C_y)) \vert \vert_1,
\end{aligned}
\end{equation}
where $\mathcal{F}^{k}$ indicates the feature map obtained by the $k$-th convolution within the VGG  network~\cite{simonyan2014very}, $W_{i,j}$ and $H_{i,j}$ are the dimensions of the respective feature maps within the VGG network.
Similarly, we have another loss for the generated image $x'$, which can be formulated as,
\begin{equation}
\begin{aligned}
 \mathcal{L}_{vgg}(x') = \frac{1}{W_{i,j}H_{i,j}}\sum_{w=1}^{W_{i,j}}\sum_{h=1}^{H_{i,j}} \vert \vert \mathcal{F}^{k} (x)-\mathcal{F}^{k}(G(y, C_x)) \vert \vert_1.
\end{aligned}
\end{equation}
Thus, the final perceptual loss is the sum of both items, i.e., $\mathcal{L}_{vgg} {=} \mathcal{L}_{vgg}(y') {+}\mathcal{L}_{vgg}(x')$.

\noindent \textbf{Total Variation Loss.}
Usually, the images synthesized by GAN models have many unfavorable artifacts, which deteriorate the visualization and recognition performance.
We impose the Total Variation (TV) loss~\cite{johnson2016perceptual} on the final synthesized image $y'$ to alleviate this issue,
\begin{equation}\small
\begin{aligned}
\mathcal{L}_{tv}(y') =  \sum_{c=1}^{C} \sum_{w,h=1}^{W,H} & \left| y'(w+1, h, c) - y'(w, h, c)  \right|  \\ + 
&\left| y'(w, h+1, c) - y'(w, h, c) \right|,
\end{aligned}
\label{eqn:tv}
\end{equation}
where $W$ and $H$ represent the width and height of the generated image $y'$.
Similarly, we have another loss for the generated image $x'$ and the final total variation loss is the sum of both.

\noindent \textbf{Overall Loss.} 
The total optimization loss is a weighted sum of the above losses. Generator $G$ and discriminator $D$  are trained in an end-to-end fashion to optimize the following min-max function,
\begin{equation}
\begin{aligned}
G^*=\arg \min\limits_G \max\limits_D & (\mathcal{L}_{adv} + \lambda_{color} \mathcal{L}_{color} + \lambda_{cyc} \mathcal{L}_{cyc} + \\
& \lambda_{con} \mathcal{L}_{con} + \lambda_{vgg} \mathcal{L}_{vgg} + \lambda_{tv} \mathcal{L}_{tv}),
\end{aligned}
\label{eqn:overallloss}
\end{equation}
where $\lambda_{color}$, $\lambda_{cyc}$, $\lambda_{con}$, $\lambda_{vgg}$ and $\lambda_{tv}$ are five hyper-parameters controlling the relative importance of these six losses. 
Solving this min-max problem enables our model to generate the target images guided by controllable structures in a photo-realistic manner.

\subsection{Implementation Details}

\noindent \textbf{Network Architecture.}
We adopt our generator architecture $G$ from~\cite{johnson2016perceptual}, which has shown effective in many applications such as unsupervised image translation~\cite{zhu2017unpaired} and neural style transfer~\cite{johnson2016perceptual}.
We use 9 residual blocks for both $64{\times}64$ and $256{\times}256$ image resolutions.
The last layer of the generator is the Tanh activation function.
For the discriminator $D$, we adopt $70{\times}70$ PatchGAN proposed in \cite{isola2017image}. PatchGAN tries to decide if any $70{\times}70$ patch in an image is real or fake.
The final layer of discriminators employs the Sigmoid activation function to produce a 1-dimensional output.
Therefore, we are averaging all responses to provide the ultimate output of the discriminator.

\noindent \textbf{Optimization Details.}
We observe that the proposed controllable structure guided discriminator achieves promising generation results.
However, to further improve the image quality, we use the scheme of training a dual-discriminator instead of one discriminator as a more stable way to improve the capacity of discriminators similar to Nguyen et al.~\cite{nguyen2017dual}, which have demonstrated that they improve the ability of discriminator to generate more photo-realistic images.
To be more specific, dual-discriminator architecture can better approximate optimal discriminator. 
If one of the discriminators is trained to be far superior over the generators, the generators can still receive instructive gradients from the other one.
In addition to the proposed controllable structure guided discriminator, we use a traditional one, which takes the input image and the generated image  as input.
Both discriminators have the same network architecture structure.

We follow the standard optimization method in~\cite{goodfellow2014generative,isola2017image} to optimize the proposed GAN model, i.e., one gradient descent step on discriminators and generator alternately.
We first train generator $G$ with discriminators fixed, and then train discriminators with generator $G$ fixed.
In addition, as suggested in~\cite{goodfellow2014generative}, we train to maximize $\log D([x, C_y], y')$ rather than $\log (1 {-} D([x, C_y], y'))$.
Moreover, in order to slow down the rate of $D$ relative to $G$ we divide the objective function by 2 while optimizing $D$. 
The proposed GAN model is trained in an end-to-end fashion.
We employ the Adam~\cite{kingma2014adam} optimizer with momentum terms  $\beta_1{=}0.5$ and $\beta_2{=}0.999$ as our solver.
The initial learning rate for Adam is 0.0002.

We follow~\cite{ma2017pose} and exploit OpenPose \cite{simon2017hand} to detect the ground-truth hand skeletons as training data for the hand gesture-to-gesture translation task. 
We then connect the 21 keypoints (hand joints) detected by OpenPose to obtain the hand skeleton.
The hand skeleton image visually contains richer hand structure information than the hand keypoint image.
In hand skeleton images, the hand joints are connected by the lines with a width of 4 and with white color.
In addition, we follow~\cite{regmi2018cross} and use RefineNet~\cite{lin2017refinenet} to generate the ground-truth semantic maps as training data for the cross-view image translation task.

\subsection{Fr\'echet ResNet Distance}
We also propose a novel evaluation metric to measure the image quality of the generated images by GAN models, i.e., Fr\'echet ResNet Distance (FRD).
FRD provides an alternative method to quantify the quality of synthesis and is similar to the FID proposed by \cite{heusel2017gans}.
FID is a measure of similarity between two datasets of images. 
The authors have shown that the FID is more robust to noise than Inception Score~(IS) and correlates well with the human judgment of visual quality~\cite{heusel2017gans}.
To calculate FID between two image domains $y$ and $y'$, they first embed both into a feature space $F$ given by an Inception model.
Then viewing the feature space as a continuous multivariate Gaussian as suggested in \cite{heusel2017gans}, Fr\'echet distance between the two Gaussians is used to quantify the quality of the data. The definition of FID is:
\begin{equation}
\begin{aligned}
\mathrm{FID}(y, y') = \lVert \mu_y - \mu_{y'} \lVert_2^2 + \mathrm{Tr}(\textstyle{\sum_y + \sum_{y'}} - 2 (\textstyle{\sum_y \sum_{y'}})^{\frac{1}{2}}),
\end{aligned}
\label{eqn:fid}
\end{equation}
where $( \mu_y, \textstyle{\sum_y})$ and $(\mu_{y'}, \textstyle{\sum_{y'}})$ are the mean and the co-variance of the data distribution and model distribution, respectively.
Note that we regard the images in $y'$ and $y$ as two wholes respectively, and then calculate the Fr\'echet distance between $y'$ and $y$ for calculating FID.

Unlike FID, which calculates the distance between two distributions, the proposed FRD is inspired by the feature matching method~\cite{tang2016novel}, and separately calculates the Fr\'echet distance between each generated image and the corresponding real image from a semantic point of view.
In this way, images from the two domains do not affect each other when computing the Fr\'echet distance. 
Moreover, for FID the number of samples should be greater than the dimension of the coding layer, while the proposed FRD does not have this limitation. 
We denote $y_i$ and $y'_i$ as images in the domain $y$ and $y'$, respectively.
For calculating FRD, we first embed both images $y_i$ and $y'_i$ into a feature space $F$ with 1,000 dimensions given by a ResNet50 pretrained model.
We then calculate the Fr\'echet distance between two feature maps $f(y_i)$ and $f(y'_i)$.
The Fr\'echet distance $F(f(y_i),f(y'_i))$ is defined as the infimum over all reparameterizations $\alpha$ and $\beta$ of $[0, 1]$ of the maximum over all $t \in [0, 1]$ of the distance in $F$ between $f(y_i)(\alpha(t))$ and $f(y'_i)(\beta(t))$, where $\alpha$ and $\beta$ are continuous, non-decreasing surjections of the range $[0,1]$.
The proposed FRD is a measure of similarity between the feature vector of the real image $f(y_i)$  and the feature vector of the generated image $f(y'_i)$ by calculating the Fr\'echet distance between them. The Fr\'echet distance is defined as the minimum cord-length sufficient to join a point traveling forward along $f(y'_i)$ and one traveling forward along $f(y_i)$, although the rate of travel for each point may not necessarily be uniform. 
Thus, the definition of FRD between two image domains $y$ and $y'$ is:
\begin{equation}
\begin{aligned}
\mathrm{FRD}(y, y') = \frac{1}{N} \sum_{i=1}^N \inf\limits_{\alpha, \beta} \max\limits_{t \in\left[0, 1\right]} \left\lbrace  d \big(f(y_i) (\alpha (t) ), f(y'_i) (\beta(t)) \big) \right\rbrace, 
\end{aligned}
\label{eqn:frd}
\end{equation}
where $d$ is the distance function of $F$ and $N$ is the total number of images in $y$ and $y'$ domains.
\begin{figure*}[!htbp] \small
	\centering
	\includegraphics[width=1\linewidth]{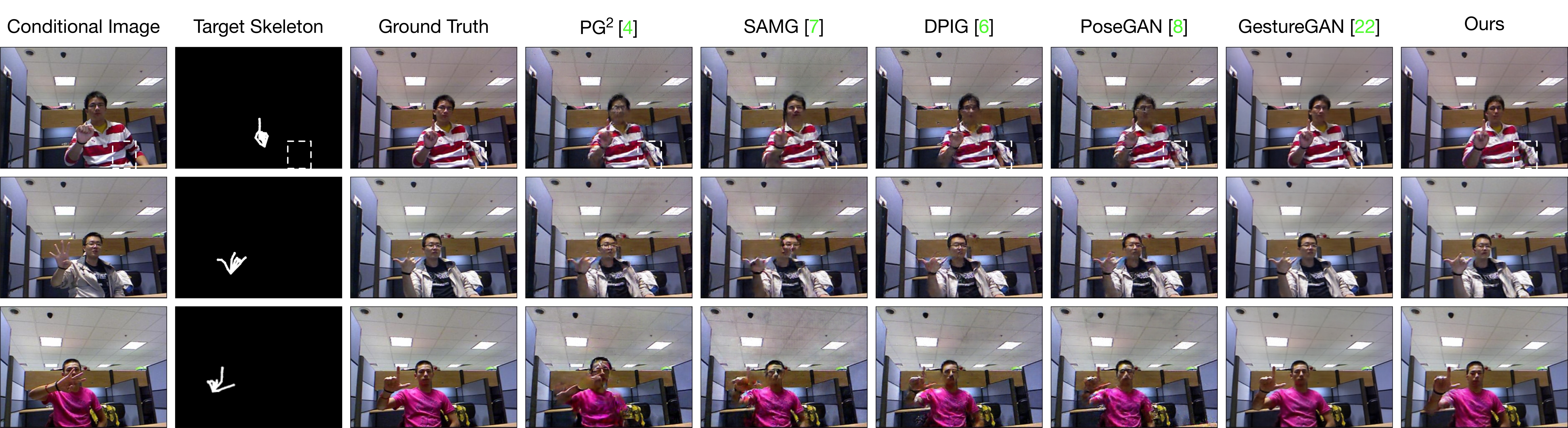}
	\caption{Different methods for hand gesture-to-gesture translation on NTU Hand Digit. 
	}
	\label{fig:comparsion_ntu}
	\vspace{-0.4cm}
\end{figure*}

\begin{figure*}[!htbp] \small
	\centering
	\includegraphics[width=1\linewidth]{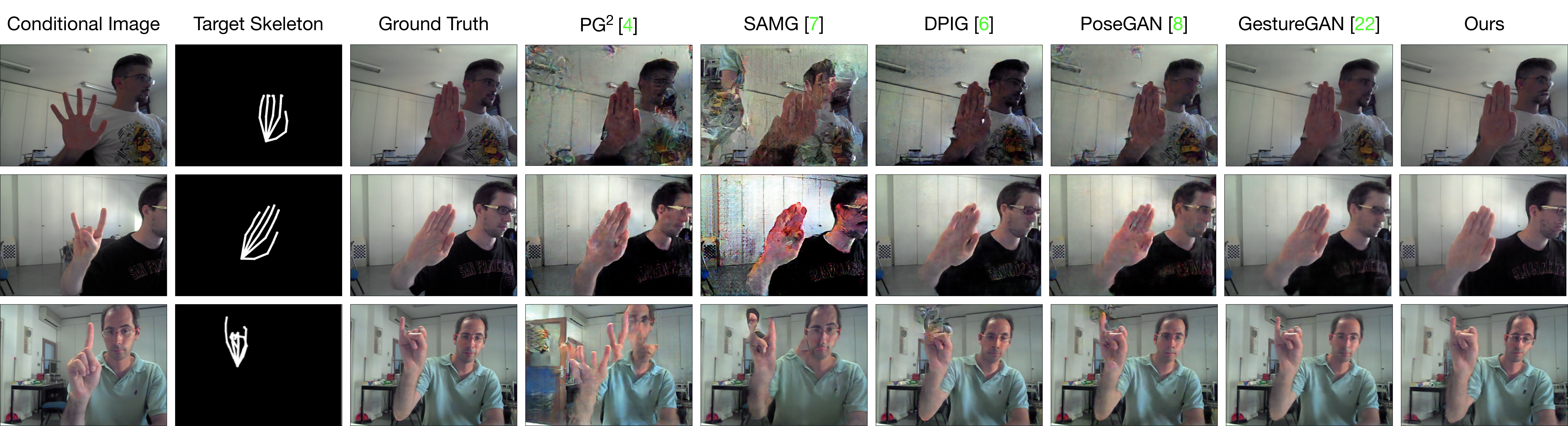}
	\caption{Different methods for hand gesture-to-gesture translation on Senz3D.
	}
	\label{fig:comparsion_3d}
	\vspace{-0.4cm}
\end{figure*}

\begin{table*}[!htbp] \small
	\centering
	\caption{Comparison results with state-of-the-art models for hand gesture-to-gesture translation on NTU Hand Digit and Senz3D. For all metrics except FID and FRD, higher is better. ($\ast$) These results are reported in \cite{tang2018gesturegan}. }
	\resizebox{\linewidth}{!}{%
		\begin{tabular}{ccccccccccc} \toprule
			\multirow{2}{*}{Method} & \multicolumn{5}{c}{NTU Hand Digit}& \multicolumn{5}{c}{Senz3D}   \\ \cmidrule(lr){2-6}\cmidrule(lr){7-11}
			& PSNR $\uparrow$ & IS  $\uparrow$& AMT $\uparrow$  & FID $\downarrow$ & FRD $\downarrow$ & PSNR $\uparrow$& IS $\uparrow$ & AMT $\uparrow$ & FID $\downarrow$ & FRD $\downarrow$  \\ \midrule			
			PG$^2$~\cite{ma2017pose}   & 28.2403$^\ast$ & 2.4152$^\ast$  & 3.5\%$^\ast$ & 24.2093$^\ast$ & 2.6319$^\ast$  & 26.5138$^\ast$ & 3.3699$^\ast$ & 2.8\%$^\ast$ & 31.7333$^\ast$ & 3.0933$^\ast$ \\ 
			SAMG~\cite{yan2017skeleton}  & 28.0185$^\ast$ & 2.4919$^\ast$ & 2.6\%$^\ast$ & 31.2841$^\ast$ & 2.7453$^\ast$ & 26.9545$^\ast$ & 3.3285$^\ast$ & 2.3\%$^\ast$ & 38.1758$^\ast$ & 3.1006$^\ast$ \\ 
			DPIG~\cite{ma2017disentangled}  & 30.6487$^\ast$ & 2.4547$^\ast$ & 7.1\%$^\ast$ &6.7661$^\ast$ & 2.6184$^\ast$ & 26.9451$^\ast$& 3.3874$^\ast$ & 6.9\%$^\ast$ & 26.2713$^\ast$ & 3.0846$^\ast$ \\  
			PoseGAN~\cite{siarohin2017deformable} & 29.5471$^\ast$ & 2.4017$^\ast$ & 9.3\%$^\ast$ & 9.6725$^\ast$ & 2.5846$^\ast$ & 27.3014$^\ast$ & 3.2147$^\ast$ & 8.6\%$^\ast$ & 24.6712$^\ast$ & 3.0467$^\ast$ \\ 
			GestureGAN~\cite{tang2018gesturegan} &32.6091$^\ast$ &\textbf{2.5532}$^\ast$  & 26.1\%$^\ast$  & 7.5860$^\ast$ &2.5223$^\ast$  &27.9749$^\ast$ &\textbf{3.4107}$^\ast$ & 22.6\%$^\ast$ &18.4595$^\ast$ &2.9836$^\ast$ \\ \hline
			Ours &\textbf{32.6574}& 2.3783 & \textbf{29.3\%}& \textbf{6.7493} & \textbf{1.7401} & \textbf{31.5420} & 2.2159 & \textbf{27.6\%}& \textbf{12.4465} & \textbf{2.2104} \\
			\bottomrule		
	\end{tabular}}
	\label{tab:gesture_comp}
	\vspace{-0.4cm}
\end{table*}

\section{Experiments}
\label{sec:experiment}
To explore the generality of the proposed GAN model, we evaluate the proposed model on a variety of tasks and datasets, including hand gesture-to-gesture translation and cross-view image translation.

\subsection{Hand Gesture-to-Gesture Translation}

\noindent \textbf{Datasets.} We follow GestureGAN~\cite{tang2018gesturegan} and evaluate the proposed GAN model on two hand gesture datasets, i.e., NTU Hand Digit~\cite{ren2013robust} and Creative Senz3D~\cite{memo2016head}, which include different hand gestures.
We use the hand gesture images and filter out failure cases in hand estimation for both training and testing sets.
1) NTU Hand Digit~\cite{ren2013robust} contains 10 hand gestures (e.g., decimal digits from 0 to 9) color images
and depth maps collected with a Kinect sensor under cluttered backgrounds.
We randomly select 84,636 pairs, each of which is comprised of two images of the same person but different gestures.
9,600 pairs are randomly selected for the testing subset and the rest of 75,036 pairs as the training set.
2) Creative Senz3D~\cite{memo2016head} includes static hand gestures performed by 4 people, each performing~11 different gestures repeated 30 times each in
the front of a Creative Senz3D camera.
We randomly select 12,800 pairs and 135,504 pairs as the testing and training set, each pair is composed of two images of the same person but different gestures.

\noindent\textbf{Parameter Settings.}
For both datasets, we do left-right flip and random crops for data augmentation.
For optimization, models are trained with a batch size of 4 for 20 epochs on both datasets.
At inference time, we follow the same settings of PG$^2$~\cite{ma2017pose} to randomly select the target keypoint or skeleton.

\noindent\textbf{Evaluation Metrics.}
Following GestureGAN~\cite{tang2018gesturegan}, we use Peak Signal-to-Noise Ratio (PSNR), Inception Score (IS), Fr\'echet Inception Distance (FID), and the proposed FRD to evaluate the quality of generated images.

\noindent \textbf{State-of-the-Art Comparisons.} We compare the proposed model with the most related works,
i.e., PG$^2$~\cite{ma2017pose}, SAMG~\cite{yan2017skeleton}, PoseGAN~\cite{siarohin2017deformable}, DPIG~\cite{ma2017disentangled} and GestureGAN~\cite{tang2018gesturegan}.
PG$^2$ and DPIG try to generate a person image with different poses based on conditional person images and target keypoints.
SAMG and PoseGAN explicitly employ human skeleton information to generate person images.  
Note that SAMG adopts a CGAN to generate motion sequences based on appearance information and skeleton information by exploiting frame-level smoothness.
We re-implemented this model to generate a single frame for a fair comparison. 
These methods are paired image-to-image models and comparison results are shown in Fig.~\ref{fig:comparsion_ntu} and~\ref{fig:comparsion_3d}.
As we can see in both figures, the proposed model consistently produces sharper images with convincing details compared with other baselines on both datasets.
We also note that the proposed GAN model is more robust than existing methods as shown in the first row of Fig.~\ref{fig:comparsion_ntu}.
Existing methods are easy to overfit since they generate the dropping arm as shown in the white dotted box while the proposed model failed to generate it.
It is hard to generate the dropping arm since no guidance has been provided to generate it, while exiting methods just simply memorize the blocks from training
images to generate new ones rather than to learn the representations between different images.

Moreover, we also provide quantitative results in Table~\ref{tab:gesture_comp}, and we can see that the proposed GAN model  produces more photo-realistic results that other baselines on all metrics expect IS.
This phenomenon can also be observed in PG$^2$~\cite{ma2017pose}, GestureGAN~\cite{tang2018gesturegan}, and other super-resolution work such as~\cite{johnson2016perceptual}, i.e., sharper results have a lower IS.
Finally, we also show some results of the arbitrary hand gesture-to-gesture translation on NTU Hand Digit dataset in Fig.~\ref{fig:hand_arbitrary}.
Given a single image and several hand skeletons, the proposed model can generate the corresponding hand gestures. 

\noindent \textbf{User Study.} 
We follow the same settings as in \cite{isola2017image} to perform an Amazon Mechanical Turk (AMT) perceptual study and gather data from 50 participants per algorithm we tested. Specifically, participants were presented a sequence of pairs of images, a `real' image and a `fake' image (generated by our algorithm or a baseline), and asked to click on the image they thought was real. The first 10 images of each session were practice and feedback was given. The remaining 40 images were used to assess the rate at which each algorithm fooled participants. Each session only tested a single algorithm, and participants were only allowed to complete a single session.
The results on NTU Hand Digit and Senz3D datasets compared with the baseline models are shown in Table \ref{tab:gesture_comp}.
We observe that the proposed model consistently achieves the best performance compared with these baselines.  

\begin{figure*}[!tbp] \small
	\centering
	\includegraphics[width=1\linewidth]{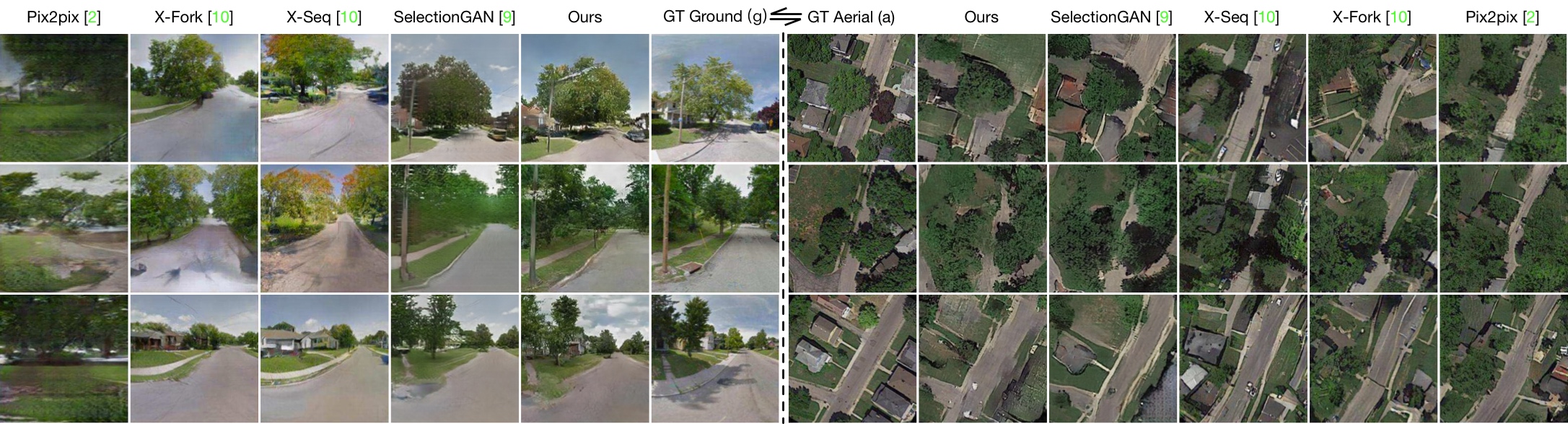}
	\caption{Different methods for cross-view image translation in $256{\times}256$ resolution on Dayton. 
	}
	\label{fig:day256}
	\vspace{-0.4cm}
\end{figure*}

\begin{figure}[!tbp] \small
	\centering
	\includegraphics[width=1\linewidth]{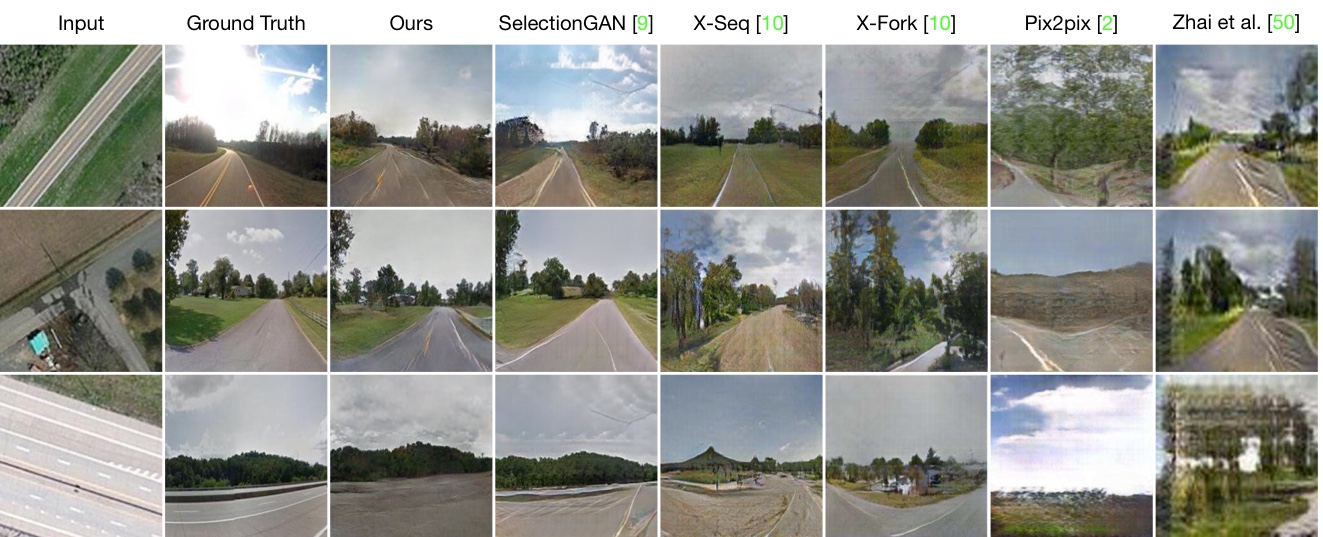}
	\caption{Different methods for cross-view image translation in $256{\times}256$ resolution on CVUSA.
	}
	\label{fig:cvusa_comparsion}
	\vspace{-0.4cm}
\end{figure}

\noindent \textbf{FID v.s. FRD.} We also compare the performance between FID and the proposed FRD metric.
The results are shown in Table~\ref{tab:gesture_comp} and we can observe that FRD is more consistent with the human judgment, i.e., the AMT score,  than the FID metric.
Moreover, we observe that the difference in FRD between GestureGAN and the other methods is not as obvious as in the results from the user study, i.e., the AMT metric.
The reason is that FRD calculates the Fr\'echet distance between the feature maps extracted from the real image and the generated image using CNNs which are trained with semantic labels. 
Thus, these feature maps are employed to reflect the semantic distance between the images. 
The semantic distance between the images is not very large considering they are all hands. 
On the contrary, the user study measures the generation quality from a perceptual level. 
The difference on the perceptual level is more obvious than on the semantic level, i.e., the generated images with small artifacts show minor differences on the feature level, while are being judged with a significant difference from the real images by humans. 

\subsection{Cross-View Image Translation}

\noindent \textbf{Datasets.}
We follow~\cite{regmi2018cross} and conduct the experiments on two public datasets: 1) For Dayton~\cite{vo2016localizing}, following the same setting of~\cite{regmi2018cross}, we select 76,048 images and create a train/test split of 55,000/21,048 pairs. 
The images in the original dataset have $354{\times}354$ resolution. 
We resize them to $256{\times}256$.
2) CVUSA \cite{workman2015wide} consists of 35,532/8,884 image pairs in train/test split. 
Following~\cite{regmi2018cross}, the aerial images are center-cropped to $224{\times}224$ and resized to $256{\times}256$. 
For the ground-level images and corresponding segmentation maps, we take the first quarter of both and resize them to $256{\times}256$.

\noindent\textbf{Parameter Settings.}
We follow~\cite{regmi2018cross} and all images are scaled to 256$\times$256, and we enabled random crops for data augmentation. 
The low-resolution experiments on Dayton are carried out for 100 epochs with a batch size of 16, whereas the high-resolution experiments for this dataset are trained for 35 epochs with a batch size of~4. 
For CVUSA, we follow the same setup as in~\cite{regmi2018cross} and train our network for 30 epochs with a batch size of 4. 

\begin{figure}[!tbp] \small
	\centering
	\includegraphics[width=1\linewidth]{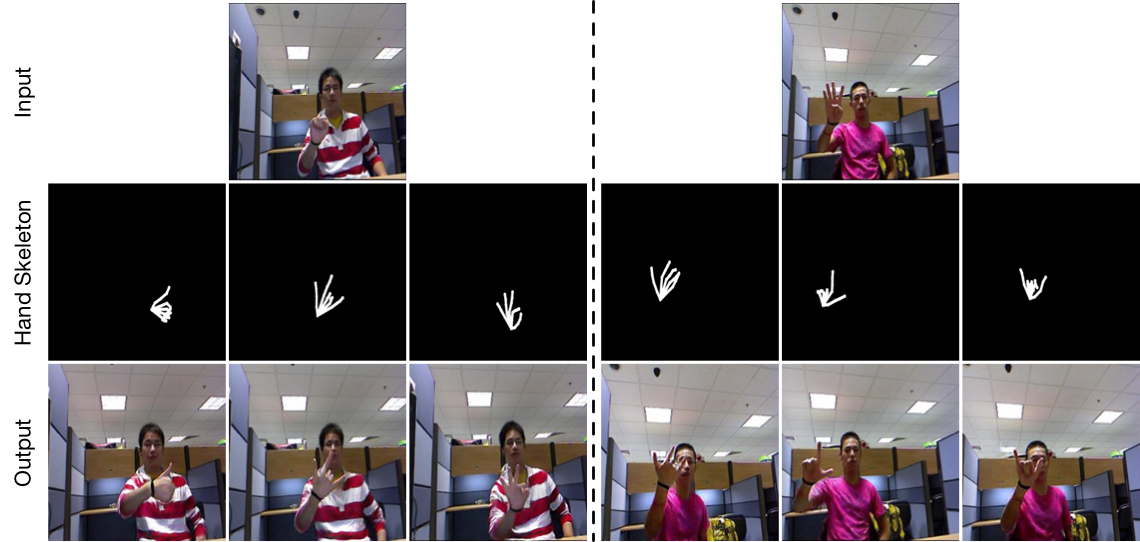}
	\caption{Arbitrary hand gesture-to-gesture translation of our model.
	}
	\label{fig:hand_arbitrary}
	\vspace{-0.4cm}
\end{figure}

\noindent\textbf{Evaluation Metrics.}
We follow \cite{regmi2018cross} and use Inception Score (IS), top-k prediction accuracy, KL score, Structural-Similarity (SSIM), PSNR, and Sharpness Difference (SD) for the quantitative analysis.
Moreover, we employ LPIPS~\cite{zhang2018unreasonable} to evaluate the quality of the generated images. LPIPS uses pretrained deep models to evaluate the
similarity, which highly agrees well with humans' perception. Specifically, we use the default pretrained AlexNet provided by the authors~\cite{zhang2018unreasonable} to calculate the LPIPS metric.

\begin{figure}[!tbp] \small
	\centering
	\includegraphics[width=1\linewidth]{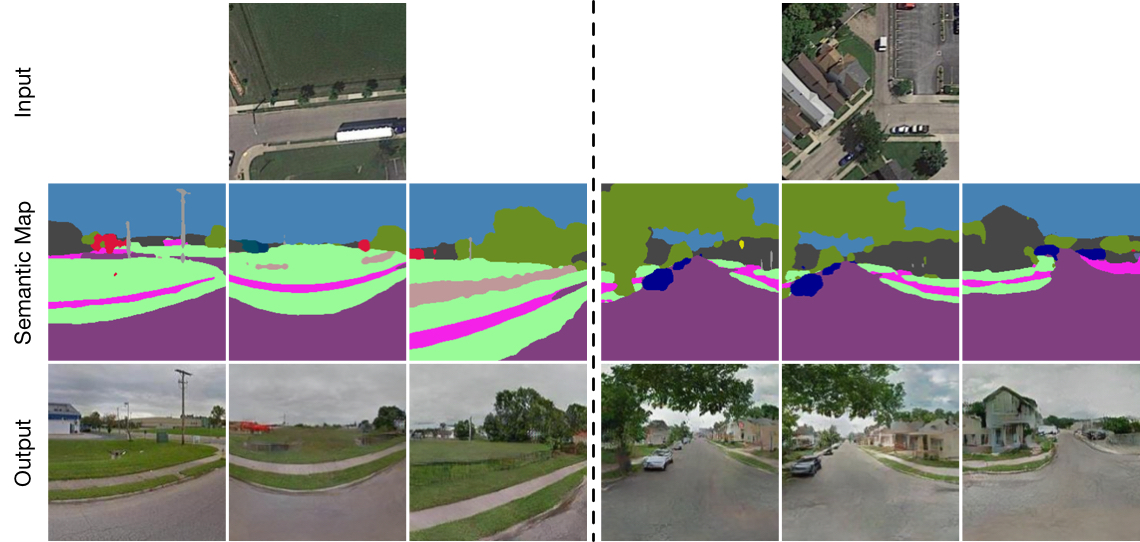}
	\caption{Arbitrary cross-view image translation of our model.
	}
	\label{fig:arbitrary}
	\vspace{-0.4cm}
\end{figure}

\textbf{\begin{table*}[!h] \small
	\centering
	\caption{Quantitative evaluation of Dayton in $64{\times}64$ resolution. For all metrics except KL score, higher is better. 
	($\ast$, $\dagger$) These results are reported in~\cite{regmi2018cross} and \cite{tang2019multi}, respectively.}
	\resizebox{1\linewidth}{!}{%
		\begin{tabular}{cccccccccccccc} \toprule
			Dir. & \multirow{2}{*}{Method} & \multicolumn{4}{c}{Accuracy (\%) $\uparrow$} & \multicolumn{3}{c}{Inception Score $\uparrow$} & \multirow{2}{*}{SSIM $\uparrow$} & \multirow{2}{*}{PSNR $\uparrow$} & \multirow{2}{*}{SD $\uparrow$} & \multirow{2}{*}{KL $\downarrow$}  \\ \cmidrule(lr){3-6} \cmidrule(lr){7-9} 
			$\leftrightarrows$ & &\multicolumn{2}{c}{Top-1} & \multicolumn{2}{c}{Top-5} & all & Top-1 & Top-5 \\ \hline
			\multirow{6}{*}{a2g}
			& Pix2pix \cite{isola2017image} & 7.90$^\ast$ & 15.33$^\ast$ & 27.61$^\ast$ & 39.07$^\ast$ & 1.8029$^\ast$ & 1.5014$^\ast$ & 1.9300$^\ast$ &0.4808$^\ast$ & 19.4919$^\ast$ & 16.4489$^\ast$ & 6.29 $\pm$ 0.80$^\ast$ \\
			& X-Fork \cite{regmi2018cross} & 16.63$^\ast$ & 34.73$^\ast$ & 46.35$^\ast$ & 70.01$^\ast$ & 1.9600$^\ast$ & 1.5908$^\ast$ & 2.0348$^\ast$ &0.4921$^\ast$ & 19.6273$^\ast$ & 16.4928$^\ast$ & 3.42 $\pm$ 0.72$^\ast$ \\
			& X-Seq \cite{regmi2018cross}             & 4.83$^\ast$ & 5.56$^\ast$ & 19.55$^\ast$ & 24.96$^\ast$ & 1.8503$^\ast$ & 1.4850$^\ast$ & 1.9623$^\ast$ &0.5171$^\ast$ &20.1049$^\ast$ &16.6836$^\ast$ & 6.22 $\pm$ 0.87$^\ast$ \\
			& SelectionGAN~\cite{tang2019multi} & 45.37$^\dagger$ & 79.00$^\dagger$ & 83.48$^\dagger$ & 97.74$^\dagger$ & \textbf{2.1606}$^\dagger$ & 1.7213$^\dagger$ & \textbf{2.1323}$^\dagger$ & \textbf{0.6865}$^\dagger$ & \textbf{24.6143}$^\dagger$ & \textbf{18.2374}$^\dagger$ & 1.70 $\pm$ 0.45$^\dagger$ \\ 
			& Ours & \textbf{49.86} & \textbf{84.41} & \textbf{86.14} & \textbf{99.61} & 2.1059 & \textbf{1.7342} & 2.0737 &0.6754&24.2814&18.1361& \textbf{1.54 $\pm$ 0.39} \\ \cmidrule(lr){2-13} 
			& Real Data & - & - & - & - & 2.3534$^\dagger$ & 1.8135$^\dagger$ & 2.3250$^\dagger$ & - & - & - & - \\
			\hline
			\multirow{6}{*}{g2a} & Pix2pix~\cite{isola2017image} & 1.65$^\ast$ & 2.24$^\ast$ & 7.49$^\ast$ & 12.68$^\ast$ & 1.7970$^\ast$ & 1.3029$^\ast$ & 1.6101$^\ast$ &0.3675$^\ast$&20.5135$^\ast$&14.7813$^\ast$& 6.39 $\pm$ 0.90$^\ast$ \\
			& X-Fork \cite{regmi2018cross}& 4.00$^\ast$ & 16.41$^\ast$ & 15.42$^\ast$ & 35.82$^\ast$ & 1.8557$^\ast$ & 1.3162$^\ast$ & 1.6521$^\ast$ &0.3682$^\ast$&20.6933$^\ast$&14.7984$^\ast$ & 4.45 $\pm$ 0.84$^\ast$  \\
			& X-Seq \cite{regmi2018cross}& 1.55$^\ast$ & 2.99$^\ast$ & 6.27$^\ast$ & 8.96$^\ast$ & 1.7854$^\ast$ & 1.3189$^\ast$ & 1.6219$^\ast$ &0.3663$^\ast$ & 20.4239$^\ast$ & 14.7657$^\ast$ & 7.20 $\pm$ 0.92$^\ast$ \\
			& SelectionGAN~\cite{tang2019multi} & 14.12$^\dagger$ & \textbf{51.81}$^\dagger$ & 39.45$^\dagger$ & 74.70$^\dagger$ & \textbf{2.1571}$^\dagger$ & \textbf{1.4441}$^\dagger$ & \textbf{2.0828}$^\dagger$ &\textbf{0.5118}$^\dagger$&23.2657$^\dagger$&16.2894$^\dagger$& 2.25 $\pm$ 0.56$^\dagger$ 
			\\ 
			& Ours & \textbf{16.65} & 44.83 & \textbf{44.03} & \textbf{77.01} & 2.0802 & 1.4360 & 2.0628 &0.5064& \textbf{23.3632}&\textbf{16.4788}& \textbf{2.16 $\pm$ 0.59} \\ \cmidrule(lr){2-13} 
			& Real Data & - & - & - & - & 2.3015$^\dagger$ & 1.5056$^\dagger$ & 2.2095$^\dagger$ & - & - & - & - \\
			\bottomrule		
	\end{tabular}}
	\label{tab:dayton64}
	\vspace{-0.4cm}
\end{table*}}

\begin{table*}[!h] \small
	\centering
	\caption{Quantitative evaluation of Dayton in $256{\times}256$ resolution. For all metrics except KL score, higher is better. 
	($\ast$, $\dagger$) These results are reported in~\cite{regmi2018cross} and \cite{tang2019multi}, respectively.}
	\resizebox{1\linewidth}{!}{%
		\begin{tabular}{cccccccccccccc} \toprule
			Dir. & \multirow{2}{*}{Method} & \multicolumn{4}{c}{Accuracy (\%) $\uparrow$} & \multicolumn{3}{c}{Inception Score $\uparrow$} & \multirow{2}{*}{SSIM $\uparrow$} & \multirow{2}{*}{PSNR $\uparrow$} & \multirow{2}{*}{SD $\uparrow$} & \multirow{2}{*}{KL $\downarrow$}  \\ \cmidrule(lr){3-6} \cmidrule(lr){7-9} 
			$\leftrightarrows$ & &\multicolumn{2}{c}{Top-1} & \multicolumn{2}{c}{Top-5} & all & Top-1 & Top-5 \\ \hline
			\multirow{6}{*}{a2g}
			& Pix2pix \cite{isola2017image}          &6.80$^\ast$ &9.15$^\ast$ &23.55$^\ast$&27.00$^\ast$& 2.8515$^\ast$&1.9342$^\ast$&2.9083$^\ast$ & 0.4180$^\ast$ &17.6291$^\ast$&19.2821$^\ast$& 38.26 $\pm$ 1.88$^\ast$ \\
			& X-Fork \cite{regmi2018cross}           &30.00$^\ast$&48.68$^\ast$&61.57$^\ast$&78.84$^\ast$& 3.0720$^\ast$&2.2402$^\ast$&3.0932$^\ast$ &0.4963$^\ast$&19.8928$^\ast$&19.4533$^\ast$  &6.00 $\pm$ 1.28$^\ast$ \\
			& X-Seq \cite{regmi2018cross}               & 30.16$^\ast$&49.85$^\ast$&62.59$^\ast$&80.70$^\ast$& 2.7384$^\ast$&2.1304$^\ast$&2.7674$^\ast$ &0.5031$^\ast$ &20.2803$^\ast$ &19.5258$^\ast$ & 5.93 $\pm$ 1.32$^\ast$ \\
			& SelectionGAN~\cite{tang2019multi} & 42.11$^\dagger$ & 68.12$^\dagger$ & 77.74$^\dagger$ & 92.89$^\dagger$ & 3.0613$^\dagger$ & 2.2707$^\dagger$ & 3.1336$^\dagger$ & \textbf{0.5938}$^\dagger$ & \textbf{23.8874}$^\dagger$ & \textbf{20.0174}$^\dagger$ & 2.74 $\pm$ 0.86$^\dagger$ \\ & Ours & \textbf{49.12} & \textbf{80.43} & \textbf{81.20} & \textbf{94.87} & \textbf{3.3210} & \textbf{2.3494} & \textbf{3.3522} & 0.5633 & 23.3515 & 19.7692 & \textbf{2.17 $\pm$ 0.77} \\ \cmidrule(lr){2-13} 
			& Real Data  & - & - & - & - & 3.8319$^\dagger$ & 2.5753$^\dagger$ & 3.9222$^\dagger$ & - & - & - & -\\ \hline
			\multirow{6}{*}{g2a} & Pix2pix~\cite{isola2017image} &10.23$^\ast$&16.02$^\ast$&30.90$^\ast$&40.49$^\ast$& 3.5676$^\ast$&2.0325$^\ast$&2.8141$^\ast$ &0.2693$^\ast$&20.2177$^\ast$&16.9477$^\ast$ &7.88 $\pm$ 1.24$^\ast$\\
			& X-Fork \cite{regmi2018cross}      &10.54$^\ast$&15.29$^\ast$&30.76$^\ast$&37.32$^\ast$& 3.1342$^\ast$&1.8656$^\ast$&2.5599$^\ast$ &0.2763$^\ast$&20.5978$^\ast$&16.9962$^\ast$ &6.92 $\pm$ 1.15$^\ast$\\
			& X-Seq \cite{regmi2018cross}     &12.30$^\ast$&19.62$^\ast$&35.95$^\ast$&45.94$^\ast$ & \textbf{3.5849}$^\ast$&2.0489$^\ast$&2.8414$^\ast$ &0.2725$^\ast$&20.2925$^\ast$&16.9285$^\ast$ & 7.07 $\pm$ 1.19$^\ast$ \\
			& SelectionGAN~\cite{tang2019multi} & \textbf{20.66}$^\dagger$ & \textbf{33.70}$^\dagger$ & \textbf{51.01}$^\dagger$ & \textbf{63.03}$^\dagger$ & 3.2446$^\dagger$ & \textbf{2.1331}$^\dagger$ & \textbf{3.4091}$^\dagger$ & 0.3284$^\dagger$ & 21.8066$^\dagger$ & 17.3817$^\dagger$ & \textbf{3.55 $\pm$ 0.87}$^\dagger$
			\\ 
			& Ours & 17.31 & 29.40 & 43.58 & 55.27 & 3.2131 & 2.0916 & 3.3637 & \textbf{0.3357} & \textbf{22.0273} & \textbf{17.6542} & 5.17 $\pm$ 1.23 \\ \cmidrule(lr){2-13} 
			& Real Data & - & - & - & - & 3.7196$^\dagger$ & 2.3626$^\dagger$ & 3.8998$^\dagger$ & - & - & - & -\\
			\bottomrule		
	\end{tabular}}
	\label{tab:dayton}
	\vspace{-0.4cm}
\end{table*}

\noindent \textbf{State-of-the-Art Comparison.}
We compare the proposed model with five recently proposed state-of-the-art methods on the cross-view image translation task, i.e.,  Pix2pix~\cite{isola2017image}, Zhai~et al.~\cite{zhai2017predicting}, X-Fork~\cite{regmi2018cross}, X-Seq~\cite{regmi2018cross} and SelectionGAN~\cite{tang2019multi}.
The comparison results in higher resolution on the Dayton and CVUSA dataset are shown in Fig.~\ref{fig:day256} and~\ref{fig:cvusa_comparsion}, respectively.
We can see that the proposed model generates better results against other baselines in terms of detail preservation and translation visual effects. 
In addition, it can be seen that our method generates more clear details on objects/scenes such as road, trees, and clouds than SelectionGAN in the generated ground-level images (zoom-in for details in Fig.~\ref{fig:day256}). 
For the generated aerial images, we can observe that grass, trees, and house roofs are well-rendered compared to others.
Moreover, the results generated by our method are closer to the ground truth in layout and structure.

The quantitative comparison results are shown in Tables~\ref{tab:dayton64},~\ref{tab:dayton},~\ref{tab:cvusa} and~\ref{tab:lpips}.
We can observe the significant improvement of the proposed model in these tables. 
The proposed model consistently outperforms Pix2pix, Zhai~et al., X-Fork, and X-Seq on all the metrics. 
Moreover, comparing against SelectionGAN, the proposed model still achieves competitive performance on all metrics excepting SSIM, PSNR, and SD.
In most cases of the a2g direction in Tables~\ref{tab:dayton64} and \ref{tab:dayton} we achieve a slightly lower performance as compared with SelectionGAN. 
However, the proposed method consistently achieves better performance than SelectionGAN on the LPIPS metric as shown in Table~\ref{tab:lpips}, which agrees more with human judgments as indicated in~\cite{zhang2018unreasonable}.
We also report both FID and FRD results compared with the most related SelectionGAN in Tables~\ref {tab:fid} and \ref{tab:frd}.
We can see that the proposed method achieves better results than SelectionGAN in most cases.
Finally, we also note that SelectionGAN is carefully designed for the cross-view image translation task while the proposed model is a generic framework. 

\noindent \textbf{Arbitrary Cross-View Image Translation.} Existing methods such as Zhai et al.~\cite{zhai2017predicting}, Pix2pix~\cite{isola2017image}, X-Fork~\cite{regmi2018cross} and X-Seq~\cite{regmi2018cross} focus on the cross-view image translation task.
However, this task is essentially an ill-posed problem and has limited scalability and robustness in handling more than two viewpoints.
A recent work SelectionGAN~\cite{tang2019multi} extends the cross-view image translation task to a more generic task of the problem, i.e., the arbitrary cross-view image translation, in which a single input view can be translated to different target views.
For the arbitrary cross-view image translation, conditional labels are usually required since learning a one-to-many mapping is more challenging and extremely hard to optimize.
Similarly to the arbitrary hand gesture-to-gesture translation in Fig.~\ref{fig:hand_arbitrary}, we show several results of arbitrary cross-view image translation on Dayton in Fig.~\ref{fig:arbitrary}.
We believe this task has many applications such as cross-view image geo-localization.

\noindent \textbf{Network Parameter Comparisons.} We compare the overall network parameter with Pix2pix~\cite{isola2017image}, X-Fork~\cite{regmi2018cross}, X-Seq~\cite{regmi2018cross} and SelectionGAN~\cite{tang2019multi} on cross-view image translation task.
Results are shown in Table~\ref{tab:capacity}. As we can see, the proposed model achieves superior model capacity and produces better generation performance comparing with existing models. 

\subsection{Ablation Study}
We perform an ablation study in the a2g (aerial-to-ground) direction on Dayton for cross-view image translation.
Following \cite{tang2019multi}, to reduce the training time, we randomly select 1/3 samples from the whole 55,000/21,048 samples, i.e., around 18,334 samples for training and 7,017 samples for testing. 

\begin{table*}[!t] \small
	\centering
	\caption{Quantitative evaluation of CVUSA in the a2g direction. For all metrics except KL score, higher is better. ($\ast$, $\dagger$) These results are reported in~\cite{regmi2018cross} and \cite{tang2019multi}, respectively.
	}
	\resizebox{1\linewidth}{!}{%
		\begin{tabular}{cccccccccccccc} \toprule
			\multirow{2}{*}{Method}  & \multicolumn{4}{c}{Accuracy (\%) $\uparrow$}& \multicolumn{3}{c}{Inception Score $\uparrow$} & \multirow{2}{*}{SSIM $\uparrow$} & \multirow{2}{*}{PSNR $\uparrow$} & \multirow{2}{*}{SD $\uparrow$} & \multirow{2}{*}{KL $\downarrow$} \\ \cmidrule(lr){2-5} \cmidrule(lr){6-8} 
			& \multicolumn{2}{c}{Top-1} & \multicolumn{2}{c}{Top-5} & all & Top-1 & Top-5  \\ \hline
			Zhai et al.~\cite{zhai2017predicting}   &13.97$^\ast$&14.03$^\ast$&42.09$^\ast$&52.29$^\ast$ & 1.8434$^\ast$&1.5171$^\ast$ &1.8666$^\ast$ & 0.4147$^\ast$&17.4886$^\ast$&16.6184$^\ast$  & 27.43 $\pm$ 1.63$^\ast$  \\
			Pix2pix \cite{isola2017image} &7.33$^\ast$ &9.25$^\ast$ &25.81$^\ast$&32.67$^\ast$ & 3.2771$^\ast$&2.2219$^\ast$&3.4312$^\ast$ & 0.3923$^\ast$ &17.6578$^\ast$ &18.5239$^\ast$ & 59.81 $\pm$ 2.12$^\ast$  \\ 
			X-Fork \cite{regmi2018cross}           &20.58$^\ast$&31.24$^\ast$&50.51$^\ast$&63.66$^\ast$ &3.4432$^\ast$&2.5447$^\ast$&3.5567$^\ast$ & 0.4356$^\ast$ &19.0509$^\ast$ &18.6706$^\ast$ & 11.71 $\pm$ 1.55$^\ast$\\
			X-Seq \cite{regmi2018cross}             &15.98$^\ast$&24.14$^\ast$&42.91$^\ast$&54.41$^\ast$ &3.8151$^\ast$&2.6738$^\ast$&4.0077$^\ast$ & 0.4231$^\ast$ &18.8067$^\ast$ &18.4378$^\ast$ &15.52 $\pm$ 1.73$^\ast$  \\ 	SelectionGAN~\cite{tang2019multi} & 41.52$^\dagger$ & 65.51$^\dagger$ & 74.32$^\dagger$ & 89.66$^\dagger$ & 3.8074$^\dagger$ & 2.7181$^\dagger$ & 3.9197$^\dagger$ & 0.5323$^\dagger$ & \textbf{23.1466}$^\dagger$ & 19.6100$^\dagger$ & 2.96 $\pm$ 0.97$^\dagger$\\
			Ours & \textbf{45.06} & \textbf{70.04} & \textbf{78.31} & \textbf{93.47} & \textbf{3.9469} & \textbf{2.8779} & \textbf{4.0383} & \textbf{0.5366} & 22.8223 & \textbf{19.8276} & \textbf{2.60 $\pm$ 0.97} \\ \hline
			Real Data & - & - & - & - & 4.8741$^\dagger$ &  3.2959$^\dagger$ & 4.9943$^\dagger$ & - & - & - & - \\
			\bottomrule		
	\end{tabular}}
	\label{tab:cvusa}
	\vspace{-0.4cm}
\end{table*}

\begin{table}[!htbp] \small
	\centering
	\caption{LPIPS of \cite{tang2019multi} and the proposed method for cross-view image translation. For this metric, lower is better.}
	\resizebox{1\linewidth}{!}{%
		\begin{tabular}{ccccc} \toprule
			Dir. & Method & Dayton (64$\times$64) & Dayton (256$\times$256) & CVUSA  \\ \midrule
			\multirow{2}{*}{a2g} 
			& SelectionGAN~\cite{tang2019multi} & 0.1786 & 0.4996 &  0.4652 \\ 
			& Ours & \textbf{0.1712} & \textbf{0.3529}  & \textbf{0.3817}\\ \hline
			\multirow{2}{*}{g2a} 
			& SelectionGAN~\cite{tang2019multi} &0.2489 & 0.5264 &-	\\  
			& Ours & \textbf{0.2382} & \textbf{0.4527} & - \\
			\bottomrule		
	\end{tabular}}
	\label{tab:lpips}
	\vspace{-0.4cm}
\end{table}

\begin{table}[!htbp] \small
	\centering
	\caption{FID of \cite{tang2019multi} and the proposed method for cross-view image translation. For this metric, lower is better.}
	\resizebox{1\linewidth}{!}{%
		\begin{tabular}{ccccc} \toprule
			Dir. & Method & Dayton (64$\times$64) & Dayton (256$\times$256) & CVUSA  \\ \midrule
			\multirow{2}{*}{a2g} 
			& SelectionGAN~\cite{tang2019multi} & 28.4787 & 38.3498 & \textbf{43.1102} \\ 
			& Ours & \textbf{18.7225} & \textbf{35.9220} & 47.3500 \\ \hline
			\multirow{2}{*}{g2a} 
			& SelectionGAN~\cite{tang2019multi} & 60.7903 & \textbf{85.4072} & -\\  
			& Ours & \textbf{60.1969} & 88.8195 & - \\
			\bottomrule		
	\end{tabular}}
	\label{tab:fid}
	\vspace{-0.4cm}
\end{table}

\noindent\textbf{Baseline Models.} The proposed GAN model has 9 baselines (A, B, C, D, E1x, E2x, E3, E4x, F) as shown in Table~\ref{tab:ablation}. 
Baseline A uses a CycleGAN model~\cite{zhu2017unpaired} and generates $y'$ using an unpaired image $x$. 
Baseline B uses a Pix2pix structure~\cite{isola2017image}, and generates $y'$ based on $x$ using a supervised way. 
Baseline C also uses the Pix2pix structure and inputs the combination of a conditional image $x$ and the controllable structure $C_y$ to the proposed controllable structure guided generator $G$.
Baseline D uses the proposed controllable structure guided cycle upon baseline C.
Baseline E1x explores the proposed color loss in several different ways to avoid the `channel pollution' issue. 
Baseline E2x employs the proposed controllable structure guided discriminator to stabilize the optimization process.
Baseline E3 adds the proposed controllable structure guided self-content preserving loss to preserve content information.
Baseline E4x adds the perceptual loss and the Total Variation loss on the generated result $y'$.
Baseline F is our full model integrating baselines D, E16, E22, E3, and E42.
All the baseline models are trained and tested on the same data using the same configuration.

Note that each baseline in E (i.e., E1x, E2x, E3, and E4x) focuses on improving each aspect of the performance of the generated images.
More specifically, the proposed color loss aims to avoid the `channel pollution' issue and thus improve the pixel-level similarity metrics, i.e., SSIM, PSNR, and SD.
The proposed controllable structure guided discriminator tries to improve the structure accuracy since the controllable structure can provide strong supervision to the discriminator.
The proposed controllable structure guided self-content preserving loss can push the generated data distribution close to the real data distribution.
Finally, the perceptual loss and the Total Variation loss aim to improve image fidelity.

\begin{table}[!tbp] \small
	\centering
	\caption{FRD of \cite{tang2019multi} and the proposed method for cross-view image translation. For this metric, lower is better.}
	\resizebox{1\linewidth}{!}{%
		\begin{tabular}{ccccc} \toprule
			Dir. & Method & Dayton (64$\times$64) & Dayton (256$\times$256) & CVUSA  \\ \midrule
			\multirow{2}{*}{a2g} 
			& SelectionGAN~\cite{tang2019multi} & 3.3066 & 3.5060 & 3.1641 \\ 
			& Ours & \textbf{3.1658} & \textbf{3.3694} & \textbf{3.1547} \\ \hline
			\multirow{2}{*}{g2a} 
			& SelectionGAN~\cite{tang2019multi} & 3.8033 & \textbf{3.7646} & - \\  
			& Ours & \textbf{3.7078} & 3.8943 & - \\
			\bottomrule		
	\end{tabular}}
	\label{tab:frd}
	\vspace{-0.4cm}
\end{table}

\begin{table}[!tbp] \small
	\centering
	\caption{Comparison of the number of network parameters on cross-view image translation.}
	\resizebox{1\linewidth}{!}{%
		\begin{tabular}{cccccc} \toprule
Model                 & Pix2pix~\cite{isola2017image}     & X-Fork~\cite{regmi2018cross} & X-Seq~\cite{regmi2018cross}  & SelectionGAN~\cite{tang2019multi} & Ours \\ \midrule	
$G$         & 39.0820 M  & 39.2163 M  & 39.0820*2 M  &  55.4808 M & 11.3876 M \\  
$D$    & 2.7696 M    & 2.7696 M   & 2.7696*2 M & 2.7687 M & 2.7678+2.7709 M \\ \hline
Total                  &  41.8516 M  & 41.9859 M  & 83.7032 M & 58.2495 M & \textbf{16.9263 M}  \\        	
			\bottomrule		
	\end{tabular}}
	\vspace{-0.4cm}
	\label{tab:capacity}
\end{table}

\begin{figure}[!tbp] \small
	\centering
	\includegraphics[width=1\linewidth]{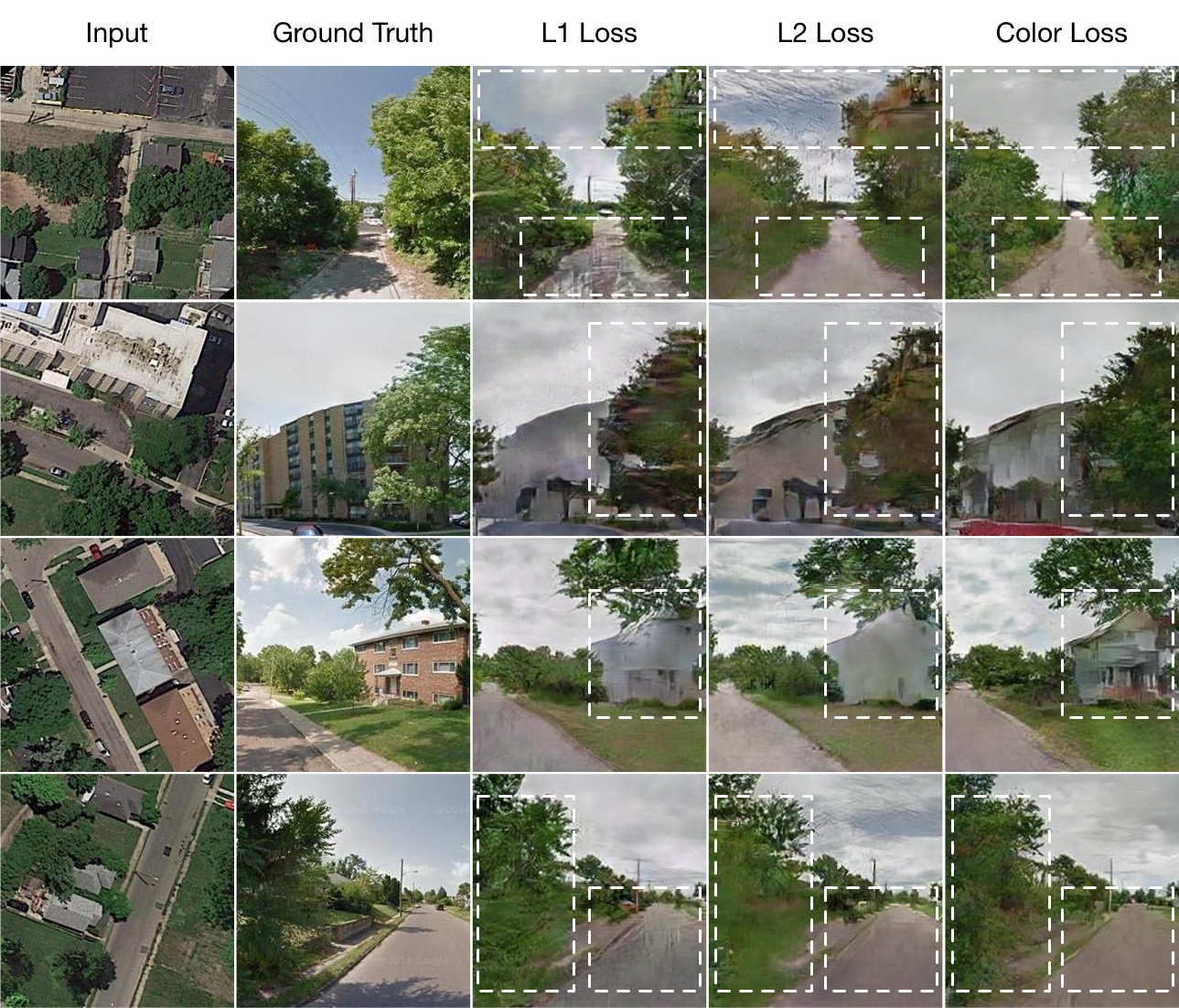}
	\caption{Comparison results of $L1$ Loss, $L2$ Loss, and the proposed Color Loss for cross-view image translation.}
	\label{fig:result_color}
	\vspace{-0.4cm}
\end{figure}

\begin{table*}[!t]\small
	\centering
	\caption{Ablation study of the proposed method on Dayton for cross-view image translation. For all evaluation metrics except KL and LPIPS, higher is better.}
	\resizebox{1\linewidth}{!}{%
		\begin{tabular}{clccccccccc} \toprule
			\multirow{2}{*}{Baseline} & \multirow{2}{*}{Experimental Setting}  & \multirow{2}{*}{SSIM $\uparrow$}  & \multirow{2}{*}{PSNR $\uparrow$} & \multirow{2}{*}{SD $\uparrow$}  &\multicolumn{4}{c}{Accuracy $\uparrow$} &  \multirow{2}{*}{KL $\downarrow$}  &  \multirow{2}{*}{LPIPS $\downarrow$}    \\   \cmidrule(lr){6-9}              
			& &   &   &  & \multicolumn{2}{c}{Top-1}  &  \multicolumn{2}{c}{Top-5} &  \\ \midrule	
			A & $x \stackrel{G} \rightarrow y'$ (Unsupervised Learning) & 0.4110 & 17.9868 & 18.5195 & 27.28 &47.47 & 52.47 & 71.63 & 8.69$\pm$1.36 & 0.5913 \\ 
			B & $x \stackrel{G} \rightarrow y'$ (Supervised Learning)     & 0.4555 & 19.6574 & 18.8870 &27.46 &46.84 &58.20 & 77.17  & 6.25$\pm$1.30 & 0.5520 \\ 
			C & $[x, C_x] \stackrel{G} \rightarrow y'$ (Controllable Structure Guided Generation) & 0.5374  & 22.8345 & 19.2075  & 39.76 & 68.44 & 72.22 & 89.85 & 3.32$\pm$1.10 & 0.4010 \\ 
			D & $[x, C_y] \stackrel{G} \rightarrow [y', C_x] \stackrel{G} \rightarrow x'$ (Controllable Structure Guided Cycle) &  0.5547 & 23.1531 & 19.6032  & 42.43 & 70.82 &75.40 & 91.16 & 2.89$\pm$1.05 & 0.3821 \\ \hline 
			
			E11 & D + Color L1 Loss on $x'$ & 0.5515 & 23.1345 & 19.6257  & 41.08 & 68.31 & 75.26 & 90.60 & 3.02$\pm$1.09 & 0.3968 \\ 
			E12 & D + L1 Loss on $y'$ & 0.5541 & 23.1492 & 19.6423 & 41.73 & 68.99 & 75.13 & 89.48 & 2.89$\pm$1.02 & 0.3835 \\
			E13 & D + L2 Loss on $y'$ & 0.5481 & 23.0939 & 19.5534 & 43.51 & 72.08 & 75.79 & 91.23 & 2.86$\pm$0.99 & 0.3913\\
			E14 & D + Color L1 Loss on $y'$ & 0.5600  & 23.3692 & 19.7018 & 44.38 & 73.21 & 75.93 & 91.69 & 2.73$\pm$0.98 & 0.3782  \\ 	
			E15 & D + Color L2 Loss on $y'$ + L1 loss on $y'$ & 0.5568 & 23.3930 & 19.6273 & 43.19 & 72.58 & 75.63 & 91.67 & 2.77$\pm$1.10 & 0.3793 \\ 
			E16 & D + Color L1 Loss on $y'$ + L1 loss on $y'$ & \textbf{0.5631} & \textbf{23.4600} & \textbf{19.7650}  & 44.97 & 73.65 & 76.28 & 92.32 & 2.70$\pm$1.08 & 0.3765 \\ \hline	
			
			E21 & D + Controllable Structure Guided Discriminator & 0.5340  & 22.8176 & 19.4404 & 43.08 & 72.80 & 74.98 & 90.89 & 3.06$\pm$1.09 & 0.4003  \\ 	
			E22 & D + Dual Discriminator &  0.5255 & 22.5405 & 19.4104  & 43.12 & 74.85 & 76.14 & 91.23 & 2.93$\pm$1.02 & 0.3937 \\ \hline 
			E3  & D + Controllable Structure Guided Self-Content Preserving Loss  & 0.5473 & 23.0475 & 19.5561 & 42.81 & 70.18 & 76.71 & 91.32 & 2.76$\pm$0.99 & 0.3877  \\ \hline 
			E41 & D + Perceptual Loss     & 0.5494 & 23.1075 & 19.5197 & 45.34 & 75.40 & 78.09 & 93.24 & 2.87$\pm$0.79 & 0.3545 \\ 
			E42 & D + Perceptual Loss + Total Variation Loss & 0.5577 & 23.0242 & 19.4943 & 44.76 & 73.96 & 77.81 & 93.69 & 2.84$\pm$0.79 & \textbf{0.3543} \\ \hline
			F & D + E16 + E22 + E3 + E42  & 0.5603 & 23.1626 &19.7455 & \textbf{46.43} & \textbf{76.94} & \textbf{79.54} & \textbf{94.33} & \textbf{2.35$\pm$0.84} & 0.3571 \\ 
			\bottomrule		
	\end{tabular}}
	\label{tab:ablation}
	\vspace{-0.4cm}
\end{table*}

\begin{table*}[!t]\small
	\centering
	\caption{The influence of $\lambda_{cyc}$ on Dayton for cross-view image translation. }
	\resizebox{0.85\linewidth}{!}{%
		\begin{tabular}{lcccccccccccc} \toprule
			\multirow{2}{*}{$\lambda_{cyc}$}  & \multirow{2}{*}{SSIM $\uparrow$}  & \multirow{2}{*}{PSNR $\uparrow$} & \multirow{2}{*}{SD $\uparrow$} & \multicolumn{3}{c}{Inception Score $\uparrow$}                                             &\multicolumn{4}{c}{Accuracy $\uparrow$} &  \multirow{2}{*}{KL $\downarrow$}  &  \multirow{2}{*}{LPIPS $\downarrow$} \\   \cmidrule(lr){5-7}  \cmidrule(lr){8-11}                   
			&    &   &  & \tht{c}{all} & \tht{c}{Top-1} & \tht{c}{Top-5} & \multicolumn{2}{c}{Top-1}  &  \multicolumn{2}{c}{Top-5} &  \\ \midrule	
			100 & 0.5383 & 23.0283 & 19.5731 & 2.9278 & 1.9960 &  2.9823 & 39.22 &67.86  & 69.55 & 88.03 & 3.96 $\pm$ 1.32 & 0.4082 \\ 
			10   & 0.5475 & 23.1264 & 19.5590 &3.2344 & 2.2321 &  3.2983 & 42.30  & 67.99 & 74.98  & 89.54 & \textbf{2.87 $\pm$ 1.01} & 0.3832  \\ 
			1     & 0.5478 & 23.1153 & 19.5158 & 3.1918 & 2.2025 & 3.2362 & 42.11  & \textbf{72.26}  & 75.37  & \textbf{91.33} & 2.88 $\pm$ 1.02 & 0.3869 \\ 
			0.1  & \textbf{0.5547} & \textbf{23.1731} & \textbf{19.6032} & \textbf{3.2823}  &\textbf{2.2401} & \textbf{3.3081} & \textbf{42.43} & 70.82 &\textbf{75.40} & 91.16 & 2.89 $\pm$ 1.05 & \textbf{0.3821} \\ \bottomrule		
	\end{tabular}}
	\label{tab:ablation_cyc}
	\vspace{-0.4cm}
\end{table*}

\begin{table}[!tt]\small
	\centering
	\caption{The influence of $\lambda_{con}$ on Dayton for cross-view image translation.}
	\resizebox{1\linewidth}{!}{%
		\begin{tabular}{cccccc} \toprule
			$\lambda_{con}$ & 0.1                    & 1                       & 5                     & 10                     & 100 \\ \midrule
			KL $\downarrow$ & 2.85$\pm$1.03 & 2.93$\pm$1.00 & 2.83$\pm$1.01 & 3.00$\pm$1.02 & \textbf{2.76$\pm$0.99} \\	\bottomrule
	\end{tabular}}
	\label{tab:ablation_id}
	\vspace{-0.4cm}
\end{table}

\begin{table}[!t]\small
	\centering
	\caption{The influence of $\lambda_{vgg}$ on Dayton for cross-view image translation.}
	\begin{tabular}{cccccc} \toprule
		$\lambda_{vgg}$ & 1            & 10        & 20        & 50        & 100 \\ \midrule
		LPIPS $\downarrow$ & 0.3812   & 0.3708 & 0.3628 & 0.3571  & \textbf{0.3545} \\	\bottomrule
	\end{tabular}
	\label{tab:ablation_vgg}
	\vspace{-0.4cm}
\end{table}

\noindent \textbf{Ablation Analysis.}
The results of the ablation study are shown in Table~\ref{tab:ablation}. 
We observe that Baseline B is better than baseline A since the ground truth image $y$ can provide strong supervised information to the generator $G$. Comparing Baseline B with Baseline C, the controllable structure guided generation improves the performance on all metrics by large margins, which confirms that the controllable structures can provide more structural information to the generator $G$. 
By using the proposed controllable structure guided cycle, Baseline D further improves over baseline C, meaning that the cycle structure indeed utilizes the controllable structure information in a more effective way, confirming our design motivation. 
Baseline E14 outperforms baselines D, E12, and E13 on SSIM, PSNR, and SD metrics showing the importance of using the proposed color loss to avoid the `channel pollution' issue.  
Visualization results of $L1$ loss, $L2$ loss and the proposed color $L1$ loss are shown in Fig.~\ref{fig:result_color}. We can see that the proposed color $L1$ loss generates more clear and visually plausible details than both $L1$ and $L2$ losses, which validates the effectiveness of the proposed color loss.
By further combining the color $L1$ loss and the $L1$ loss on the generated image~$y'$, we can further improve the performance as shown in baseline E16. 
However, replacing the color $L1$ loss with the color $L2$ loss will degrade the performance as shown by baseline E15 but the results are still better than using baseline D. 
We also use the proposed color loss on the reconstructed image $x'$ as presented in baseline E11, but it achieves the worst results. 
Comparing Baseline D with Baseline E21, the proposed controllable structure guide discriminator improves the top-1 accuracy by 0.65 and 1.98, which confirms
the importance of our design motivation. 
By further combining the controllable structure guide discriminator with the traditional discriminator in baseline E22, both top-1 and top-5 accuracies are further boosted. 
Baseline E3 outperforms D with around 0.13 gains on the KL metric, clearly demonstrating the effectiveness of the proposed controllable structure guided self-content preserving loss.  
By adding the perceptual loss and the TV loss in baseline E4x, the overall performance is further improved on LPIPS metric~\cite{zhang2018unreasonable}, which uses pretrained deep models to evaluate the similarity and highly agrees with human perception. 
Finally, we demonstrate the advantage of the proposed full model in baseline F, which integrates baseline D, E14, E22, E3, and E42. 
It is obvious that baseline F achieves the best results on both accuracy and KL score metrics.
However, we observe that baseline F achieves worse results on SSIM, PSNR, and SD compared with baseline E16, and at the same time, it achieves worse results on the LPIPS metric compared with baseline E42.
This is also observed in the  LPIPS paper~\cite{zhang2018unreasonable}, i.e., the traditional metrics (i.e., SSIM, PSNR, SD, FSIM) disagree with metrics based on deep architectures such as VGG ~\cite{simonyan2014very}.
Thus, we try to balance both metrics to reasonable results without dropping significantly the performance, and we still observe that baseline F achieves better performance on all SSIM, PSNR, SD, and LPIPS metrics than baseline D.

\noindent \textbf{Hyper-parameter Analysis.}
1) For cross-view image translation tasks, we follow \cite{isola2017image} and set $\lambda_{color}{=}100$ since $\mathcal{L}_{color}$ denotes a pixel-wise reconstruction loss.
We then follow \cite{tang2019multi} and set $\lambda_{tv}{=}1\mathrm{e}{-}6$.
In addition to $\lambda_{color}$ and $\lambda_{tv}$, we also introduce $\lambda_{cyc}$, $\lambda_{con}$, and $\lambda_{vgg}$.
Thus, we investigate the influence of $\lambda_{cyc}$, $\lambda_{con}$, $\lambda_{vgg}$ to the performance of our model. 
The results are shown in Tables~\ref{tab:ablation_cyc}, \ref{tab:ablation_id} and \ref{tab:ablation_vgg}.
In Table~\ref{tab:ablation_cyc}, when $\lambda_{cyc}$ becomes smaller, we achieve better results on most metrics. 
This means that adjusting the ratio of weighting parameters of the cycle can obtain further performance improvement.
This is different from CycleGAN~\cite{zhu2017unpaired}, which uses the same weights for both forward and backward cycle-consistency losses since CycleGAN tries to learning two mappings,
while in our model we only focus on generating photo-realistic result $y'$ and do not care about the quality of the reconstructed image $x''$.
Thus, the forward part has a lager weight than the backward part.
Moreover, we also investigate the influence of $\lambda_{con}$ and $\lambda_{vgg}$. 
The results are listed in Tables~\ref{tab:ablation_id} and~\ref{tab:ablation_vgg}.
When both $\lambda_{con}$ and $\lambda_{vgg}$ become bigger, the generator with a larger error loss dominates the training, making the whole model generating better results.
Therefore, we empirically set $\lambda_{cyc}{=}0.1$, $\lambda_{con}{=}100$, $\lambda_{vgg}{=}100$, $\lambda_{color}{=}100$ and $\lambda_{tv}{=}1\mathrm{e}{-}6$ in Eq.~\eqref{eqn:overallloss} for this task.

\begin{table}[!tbp]\small
	\centering
	\caption{The influence of $\lambda_{cyc}$ on NTU Hand Digit for hand gesture-to-gesture translation.}
	\begin{tabular}{ccccc} \toprule
		$\lambda_{cyc}$ & PSNR $\uparrow$       & IS  $\uparrow$      & FID   $\downarrow$       & FRD $\downarrow$ \\ \midrule
		0.001                 & 28.5673    & \textbf{2.4851} & 23.9935 & 2.8468 \\
		0.01                   & 28.5475    & 2.3719 & 23.8958 & 2.8991\\
		0.1                     & 28.4967      & 2.4755 & \textbf{21.6280} &  \textbf{2.7571} \\	
		1                        & 28.5370    & 2.3436  & 23.5811 &  2.8467 \\
		10                      & 28.5627     &  2.4815 & 22.5539 & 2.8401 \\
		100                    & \textbf{28.5854}    & 2.4191  & 23.5617  & 2.8080 \\ \bottomrule
	\end{tabular}
	\label{tab:ablation_ntu_cyc}
	\vspace{-0.4cm}
\end{table}

\begin{table}[!tbp]\small
	\centering
	\caption{The influence of $\lambda_{con}$ on NTU Hand Digit for hand gesture-to-gesture translation.}
	\begin{tabular}{ccccc} \toprule
		$\lambda_{con}$ & PSNR $\uparrow$        & IS $\uparrow$       & FID $\downarrow$         & FRD $\downarrow$ \\ \midrule
		0.001                 &  28.5638 &2.4335 & 20.6123 & 2.7273 \\
		0.01                   &  28.4607  &2.3665 & \textbf{19.9356}   & \textbf{2.6960} \\
		0.1                     &  \textbf{28.6696} &2.3446 & 23.2919  & 2.8326 \\	
		1                        & 28.6478 &2.3522  & 24.4331 & 2.9171 \\
		10                      & 28.6642  &2.3528 & 21.7138   & 2.8778\\
		100                    & 28.5207 &\textbf{2.4881} & 24.3938 & 2.9104 \\ \bottomrule
	\end{tabular}
	\label{tab:ablation_ntu_identity}
	\vspace{-0.4cm}
\end{table}

\begin{table}[!tbp]\small
	\centering
	\caption{The influence of $\lambda_{vgg}$ on NTU Hand Digit for hand gesture-to-gesture translation.}
	\begin{tabular}{ccccc} \toprule
		$\lambda_{vgg}$ & PSNR $\uparrow$ & IS  $\uparrow$      & FID $\downarrow$         & FRD $\downarrow$ \\ \midrule
		0.001                 & 28.4537   &2.3096  & 23.7465  & 2.6976\\
		0.01                   &  28.5580  &\textbf{2.4825}  & 23.4135   & 2.6966\\
		0.1                     &   28.5741 & 2.4684 & 23.1802   & 2.6872 \\	
		1                        &  28.5625 & 2.3182  & 20.1516  & 2.6653 \\
		10                      &  28.5486  & 2.2502  & 19.8930 & 2.6004\\
		100                    &  \textbf{28.9545}  &  2.0455  & 17.1370 & 2.4461\\
		1000                  &  28.8131  & 2.0965  &\textbf{14.1617} & \textbf{2.2135} \\ 
		10000                & 27.4805  &2.4538  & 65.1080  &3.2607 \\ \bottomrule
	\end{tabular}
	\label{tab:ablation_ntu_vgg}
	\vspace{-0.4cm}
\end{table}

\begin{table}[!tbp]\small
	\centering
	\caption{The influence of $\lambda_{color}$ on NTU Hand Digit for hand gesture-to-gesture translation.}
	\begin{tabular}{ccccc} \toprule
		$\lambda_{color}$ & PSNR $\uparrow$   & IS $\uparrow$       & FID  $\downarrow$        & FRD $\downarrow$ \\ \midrule
		100                       & 28.8131 & 2.0965 & 14.1617 & 2.2135 \\
		200                      & 29.2343 & 2.1537 & \textbf{13.4811} & 2.2421 \\
		500                      & 29.9973 & 2.1332 & 13.4823 & \textbf{2.2039}\\
		800                      & 30.4531 &2.1898 & 13.9475 & 2.2176 \\
		1000                     & 30.7087 &\textbf{2.2138}   & 15.3634  & 2.2134 \\ 
		2000                    &  31.4232 & 2.1991 & 17.1864    & 2.2872 \\ 
		5000                  & \textbf{32.3025} & 2.1022 & 28.5587   & 2.3715 \\ \bottomrule
	\end{tabular}
	\label{tab:ablation_ntu_l1}
	\vspace{-0.4cm}
\end{table}

2) For hand gesture-to-gesture translation tasks,
we first follow \cite{tang2019multi} and set $\lambda_{tv}{=}1\mathrm{e}{-}6$.
Next, we investigate the influence of $\lambda_{cyc}$, $\lambda_{con}$, $\lambda_{vgg}$, and $\lambda_{color}$ to the performance of our model. 
Results are shown in Tables~\ref{tab:ablation_ntu_cyc}, \ref{tab:ablation_ntu_identity}, \ref{tab:ablation_ntu_vgg}, and \ref{tab:ablation_ntu_l1}, respectively.
According to these tables, we empirically set $\lambda_{cyc}{=}0.1$, $\lambda_{con}{=}0.01$, $\lambda_{vgg}{=}1000$, $\lambda_{color}{=}800$, and $\lambda_{tv}{=}1\mathrm{e}{-}6$ in Eq.~\eqref{eqn:overallloss} for this task.
Moreover, we also investigate the influence of the number of cycles on this task. Results are shown in Table~\ref{tab:ablation_ntu_cycle} and we observe that the two-cycle framework achieves better results than one-cycle framework on most metrics.
\section{Conclusions}
\label{sec:conclusions}

In this paper, we focus on the challenging task of controllable image-to-image translation.
To this end, we propose a unified GAN framework, which can generate target images with different poses, sizes, structures, and locations based on a conditional image and controllable structures.
In this way, the conditional image can provide appearance information and the controllable structures can provide structure information for generating the final results.
Moreover, we also propose three novel losses to learn the mapping from the source domain to the target domain, i.e., color loss, controllable structure guided cycle-consistency loss, and controllable structure guided self-content preserving loss.
It is worth noting that the proposed color loss handles the `channel pollution' problem when back-propagating the gradients, which frequently occurs in the existing generative models.
The controllable structure guided cycle-consistency loss can reduce the dis-match between the source domain and the target domain.
The controllable structure guided self-content preserving loss aims to preserve the image content information of generated images.
In addition, we present a novel Fr\'echet ResNet Distance (FRD) metric to evaluate the quality of generated images.
Experimental results show that the proposed unified GAN framework achieves competitive performance compared with existing methods using carefully designed frameworks on two challenging generative tasks, i.e., hand gesture-to-gesture translation and cross-view image translation.
Note that the proposed GAN framework is not tuned to any specific controllable image-to-image translation tasks.

\noindent \textbf{Acknowledgments.}
This work is partially supported by National Natural Science Foundation of China (No.U1613209, 61673030), National Natural Science Foundation of Shenzhen (No.JCYJ20190808182209321), and by the Italy-China collaboration project TALENT.

\begin{table}[!tbp]\small
	\centering
	\caption{The influence of the number of cycles on NTU Hand Digit for hand gesture-to-gesture translation.}
	\begin{tabular}{ccccc} \toprule
		\# Cycle & PSNR $\uparrow$      & IS $\uparrow$        & FID $\downarrow$         & FRD $\downarrow$ \\ \midrule
		One-Cycle  & 30.4531 &\textbf{2.1898} & 13.9475 & 2.2176 \\
		Two-Cycle  &\textbf{31.4924} & 2.1493 & \textbf{11.2084} & \textbf{2.0774} \\ \bottomrule
	\end{tabular}
	\label{tab:ablation_ntu_cycle}
	\vspace{-0.4cm}
\end{table}


\footnotesize
\bibliographystyle{IEEEtran}
\bibliography{ref}

\begin{IEEEbiography}[{\includegraphics[width=1in,height=1.25in,clip,keepaspectratio]{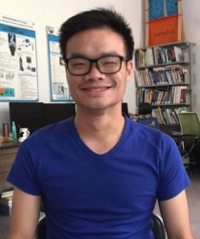}}]{Hao Tang}
is a Ph.D. candidate in the Department of Information Engineering and Computer Science at the University of Trento. He received the Master degree in computer application technology in 2016 at the School of Electronics and Computer Engineering, Peking University.
He was a visiting scholar in the Department of Engineering Science at the University of Oxford, from 2019 to 2020. His research interests are deep learning, machine learning and their applications to computer vision.
\end{IEEEbiography}

\begin{IEEEbiography}[{\includegraphics[width=1in,height=1.25in,clip,keepaspectratio]{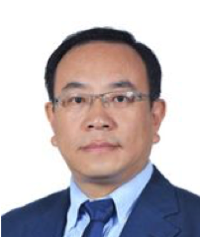}}]{Hong Liu}
received the Ph.D. degree in mechanical electronics and automation in 1996, and serves as a Full Professor in the School of EE\&CS, Peking University (PKU), China. Prof. Liu has been selected as Chinese Innovation Leading Talent supported by National High-level Talents Special Support Plan since 2013. He is also the Director of Open Lab on Human Robot Interaction, PKU. His research fields include computer vision and robotics, image processing, and pattern recognition. 
\end{IEEEbiography}

\begin{IEEEbiography}[{\includegraphics[width=1in,height=1.25in,clip,keepaspectratio]{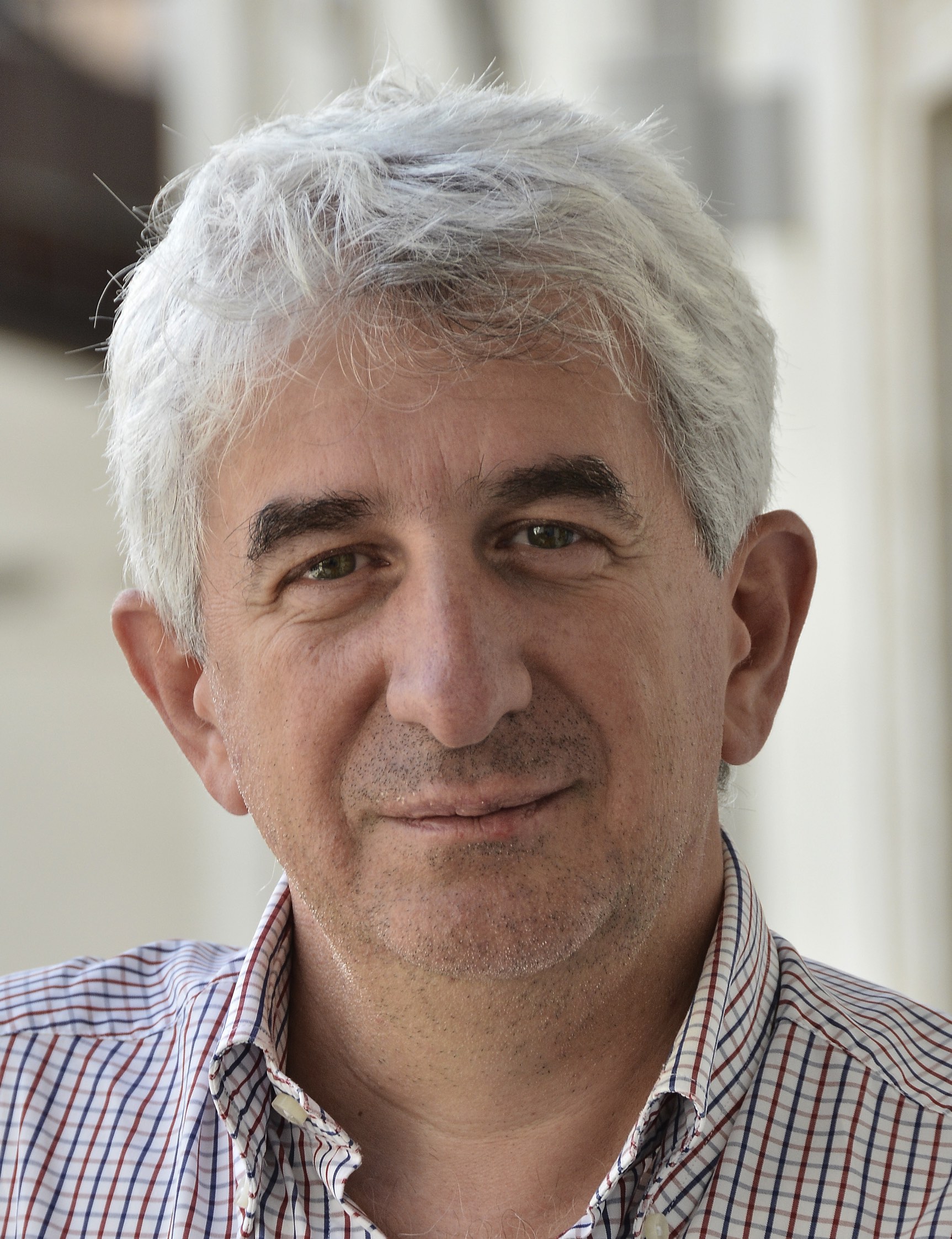}}]{Nicu Sebe} 
is Professor with the University of Trento, Italy, leading the research in the areas of multimedia information retrieval and human behavior understanding. He was the General Co-Chair of ACM Multimedia 2013, and the Program Chair of the International Conference on Image and Video Retrieval in 2007 and 2010, ACM Multimedia 2007 and 2011, ICCV 2017 and ECCV 2016. He is a Program Chair of ICPR 2020. He is a fellow of the International Association for Pattern Recognition.
\end{IEEEbiography}

%
%
%
%
%




\end{document}